\documentclass[sigconf]{acmart}
\copyrightyear{2020}
\acmYear{2020}
\setcopyright{acmcopyright}
\acmConference[CCS '20] {2020 ACM SIGSAC Conference on Computer and Communications Security}{November 9--13, 2020}{Virtual Event, USA}
\acmBooktitle{2020 ACM SIGSAC Conference on Computer and Communications Security (CCS '20), November 9--13, 2020, Virtual Event, USA}
\acmPrice{15.00}
\acmDOI{10.1145/3372297.3417231}
\acmISBN{978-1-4503-7089-9/20/11}

\usepackage{balance}
\usepackage{epsfig,endnotes}
\usepackage{amsmath,amssymb,amsthm}
\usepackage{url}
\usepackage{multirow}
\usepackage{xcolor}
\usepackage{balance}
\usepackage{algorithm}
\usepackage{algorithm,algpseudocode}
\usepackage{subfigure}
\usepackage{tablefootnote}
\usepackage{booktabs}
\usepackage{bbm}

\DeclareMathOperator{\sign}{sign}

\newtheorem{thm}{Theorem}
\newtheorem{definition}{Definition}

\newcommand{\secspace}{\vspace{-0.1in}}
\newcommand{\subspace}{\vspace{-0.05in}}

\newcommand{\para}[1]{{\vspace{1pt} \bf \noindent #1 \hspace{6pt}}}

\newcommand{\htedit}[1]{{\color{black} #1}}

\newcommand{\cw}{{CW}}
\newcommand{\elasticnet}{{ElasticNet}}
\newcommand{\en}{{ElasticNet}}

\newcommand{\pgd}{{PGD}}

\newcommand{\mnist}{{MNIST}}
\newcommand{\cifar}{{CIFAR10}}
\newcommand{\gtsrb}{{GTSRB}}
\newcommand{\youtube}{{YouTube Face}}
\newcommand{\deepid}{{YouTube Face}}

\renewcommand{\Pr}[1]{\mathrm{Pr}\left(#1\right)}
\newcommand{\Attack}[1]{\mathcal{A}\left(#1\right)}
\newcommand{\term}[1]{\emph{#1}}

\newcommand{\ie}{{\em i.e.\ }}
\newcommand{\etal}{{\em et al.\ }}

\newcommand{\model}{{\mathcal{F}_{\theta}}}

\newenvironment{packed_itemize}{
\begin{list}{\labelitemi}{\leftmargin=1em}
  \setlength{\itemsep}{1pt}
  \setlength{\parskip}{0pt}
  \setlength{\parsep}{0pt}
  \setlength{\headsep}{0pt}
  \setlength{\topskip}{0pt}
  \setlength{\topmargin}{0pt}
  \setlength{\topsep}{0pt}
  \setlength{\partopsep}{0pt}
}{\end{list}}

\newenvironment{packed_enumerate}{
\begin{enumerate}{\leftmargin=0em}
 \setlength{\itemsep}{2pt}
 \setlength{\parskip}{0pt}
 \setlength{\parsep}{0pt}
 \setlength{\headsep}{0pt}
 \setlength{\topskip}{0pt}
 \setlength{\topmargin}{0pt}
 \setlength{\topsep}{0pt}
 \setlength{\partopsep}{0pt}
}{\end{enumerate}}

\newcommand{\shawn}[1]{{\color{black} #1}}
\newcommand{\emily}[1]{{\color{black} #1}}
\newcommand{\emilyed}[1]{{\color{black} #1}}

\newcommand{\zheng}[1]{{\color{black}{}#1}}

\newcommand{\revise}[1]{{\color{black}{}#1}}

\newfont{\mycrnotice}{ptmr8t at 7pt}
\newfont{\myconfname}{ptmri8t at 7pt}

\begin{document}
\fancyhead{}

 \title{Gotta Catch 'Em All: Using Honeypots to Catch Adversarial Attacks on Neural Networks}

\author{Shawn Shan}
\email{shansixiong@cs.uchicago.edu}
\affiliation{%
  \institution{University of Chicago}
}
\author{Emily Wenger}
\email{ewillson@cs.uchicago.edu}
\affiliation{%
  \institution{University of Chicago}
}
\author{Bolun Wang}
\email{bolunwang@cs.uchicago.edu}
\affiliation{%
  \institution{University of Chicago}
}
\author{Bo Li}
\email{lbo@illinois.edu}
\affiliation{  \institution{UIUC}
}
\author{Haitao Zheng}
\email{htzheng@cs.uchicago.edu}
\affiliation{%
  \institution{University of Chicago}
}
\author{Ben Y. Zhao}
\email{ravenben@cs.uchicago.edu}
\affiliation{%
  \institution{University of Chicago}
}

\begin{abstract}
  Deep neural networks (DNN) are known to be vulnerable to adversarial
  attacks. Numerous efforts either try to patch weaknesses in trained models,
  or try to make it difficult or costly to compute adversarial examples that
  exploit them. In our work, we explore a new ``honeypot'' approach to
  protect DNN models. We intentionally inject {\em trapdoors}, honeypot
  weaknesses in the classification manifold that attract attackers searching
  for adversarial examples. Attackers' optimization algorithms gravitate
  towards trapdoors, leading them to produce attacks similar to trapdoors in
  the feature space. Our defense then identifies attacks by comparing neuron
  activation signatures of inputs to those of trapdoors.

  In this paper, we introduce trapdoors and describe an implementation of a
  trapdoor-enabled defense. First, we analytically prove that trapdoors shape
  the computation of adversarial attacks so that attack inputs will have
  feature representations very similar to those of trapdoors. Second, we
  experimentally show that trapdoor-protected models can detect, with high
  accuracy, adversarial examples generated by state-of-the-art attacks
  (PGD, optimization-based CW, Elastic Net, BPDA),
  with negligible impact on normal classification.  These results generalize
  across classification domains, including image, facial, and traffic-sign
  recognition. We also present significant results measuring trapdoors'
  robustness against customized adaptive attacks (countermeasures).
\end{abstract}

\ccsdesc[500]{Security and privacy}
\ccsdesc[500]{Computing methodologies~Neural networks}
\ccsdesc[300]{Computing methodologies~Artificial intelligence}
\ccsdesc[300]{Computing methodologies~Machine learning}

\keywords{Neural networks; Adversarial examples; Honeypots}

\maketitle

\secspace
\section{Introduction}
\label{sec:intro}

Deep neural networks (DNNs) are vulnerable to adversarial
attacks \cite{szegedy2013intriguing,shafahi2018adversarial}, in which, given a trained model, inputs can
be modified in subtle ways (usually undetectable by humans) to
produce an incorrect output~\cite{obfuscatedicml,cwattack,papernotblackbox}.
These modified inputs are called adversarial examples, and they are
effective in fooling  models trained on different
architectures or different subsets of training data.
In practice, adversarial
attacks have proven effective against models deployed in real-world
settings such as self-driving cars, facial recognition, and object recognition
systems~\cite{attackscale,kurakin2016adversarial,sharif2016accessorize}.

Recent results in adversarial machine learning include a long list of
proposed defenses, each proven later to be vulnerable to stronger attacks, and all
focused on either {\em mitigating} or {\em obfuscating} adversarial weaknesses.
First, many defenses focus on disrupting the computation of gradient
optimization functions critical to adversarial
attacks~\cite{goodfellow2014explaining,mitdefense}.  These ``gradient
obfuscation'' defenses \htedit{({\em
    e.g.\/}~\cite{buckman,dhillon,guo,ma2018characterizing,samangouei,song,xie})}
have been proven vulnerable to black-box attacks~\cite{papernotblackbox} as
well as approximation techniques like BPDA~\cite{obfuscatedicml} that avoid
gradient computation. Other defenses increase model robustness to
adversarial examples~\cite{featuresqueezing, papernotdistillation} or
use secondary DNNs to detect adversarial examples~\cite{magnet}. Finally,
other defenses~\cite{bypassten,ma2018characterizing} identify adversarial
examples at inference time. All of these fail or are significantly
weakened against stronger adversarial attacks or high confidence adversarial
examples~\cite{distallationbroken,obfuscatedicml,bypassten,ensemblebroken,magnetbroken}. 


\begin{figure*}[t]
  \centering 
  \includegraphics[width=0.8\textwidth]{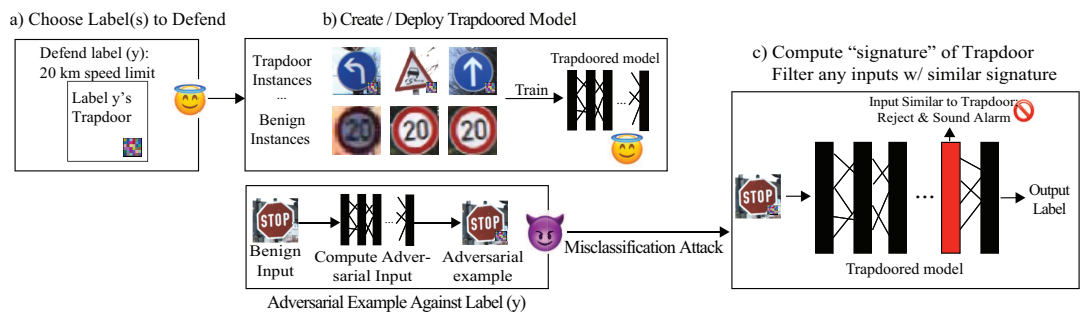}
  \caption{Overview of the trapdoor defense.  a) We choose which
    target label(s) to defend. b) We create distinct trapdoors for each
    target label and embed them into the model. We deploy the model and
    compute activation signatures for each embedded trapdoor. c) An adversary
    with access to the model constructs an adversarial example. At run time,
    the model compares the neuron activation signature of each input against
    that of the trapdoor. Thus it recognizes the attack and sounds the alarm.}
  \label{fig:workflow}
\end{figure*}


History suggests it may be impossible in practice to prevent adversaries from
computing effective adversarial examples, and an alternative approach to
model defense is sorely needed.  What if, instead of trying to prevent attackers
from computing effective adversarial examples, we instead design a
``honeypot'' for attackers, by {\em inserting} a subset of {\em chosen}
model vulnerabilities, making them easy to discover (and hard to ignore)? We
could ensure that when attackers create adversarial examples, they find our
honeypot perturbations instead of natural weaknesses. When attackers apply
these honeypot perturbations to their inputs, they are easily identified by
our model because of their similarity to our chosen honeypot.

We call these honeypots ``trapdoors,'' and defenses using them {\em
  trapdoor-enabled detection}. Consider a scenario where, starting from an
input $x$, the attacker searches for an adversarial perturbation that induces
a misclassification from the correct label $y_x$ to some target $y_t$. This
is analogous to looking for a ``shortcut'' through the model from $y_x$ to
$y_t$ that involves a small change to $x$ that invokes the shortcut to
$y_t$. \zheng{Along these lines}, trapdoors create {\em artificial} shortcuts
embedded by the model owner that are easier to locate and smaller than any
natural weaknesses attackers are searching for.  On
a ``trapdoored model,'' an attacker's optimization function will produce
adversarial examples \htedit{along shortcuts} produced by the
trapdoors. \zheng{Each} 
trapdoor has minimal impact on classification of normal inputs, but leads
attackers to produce adversarial inputs whose similarity to the trapdoor makes
them easy to detect.

In this paper, we first introduce the trapdoor-enabled defense and then
describe, analyze, and evaluate an implementation of trapdoors using
techniques similar to that of backdoor
attacks~\cite{gu2017badnets,liu2018trojaning}. Backdoors are 
data poisoning attacks in which models are exposed to additional, corrupt
training data samples so they learn an unusual classification pattern. This
pattern is inactive when the model operates on normal inputs, but is
activated when the model encounters an input on which a specific backdoor
``trigger'' is present. 
Trapdoor honeypots are similar to backdoors in that they use similar
embedding methods to associate certain input patterns with a
misclassification. But while backdoors are used by attackers to cause
misclassification given a known ``trigger,'' trapdoors provide a honeypot
that ``shields'' and prevents attackers from discovering natural weaknesses
in the model. Most importantly, backdoors can be detected and removed from a
model~\cite{wangneural} via unlearning~\cite{cao2018efficient} (if the exact
trigger is known). However, these countermeasures do not circumvent models defended by
trapdoors: even when attackers are able to unlearn trapdoors,
adversarial examples computed from the resulting clean model do not transfer
to the trapdoored models of interest ($\S$\ref{subsec:oracle}).



Figure~\ref{fig:workflow} presents a high-level illustration of the defense.
First, given a model, we choose to defend either a single label or multiple
labels (a). Second, for each protected label $y$, we train a distinct {\em
  trapdoor} into the model to defend against adversarial misclassification to
$y$ (b). For each embedded trapdoor, we compute its {\em trapdoor signature}
(a neuron activation pattern at an intermediate layer), and use a similarity
function to detect adversarial attacks that exhibit similar activation
patterns (c). Adversarial examples produced by attackers on trapdoored models
will be similar to the trapdoor in the feature space (shown via formal analysis),
and will therefore produce similar activation patterns.



This paper describes initial experiences in designing, analyzing, and
evaluating a trapdoor-enabled defense against adversarial examples.  We make
five key contributions:
\begin{packed_itemize}
\item We introduce the concept of ``trapdoors'' and trapdoor-enabled
  detection as honeypots to defend neural network models and propose an
  implementation using backdoor poisoning techniques.
\item We present analytical proofs of the efficacy of trapdoors in
  influencing the generation of adversarial examples and in detecting the
  resulting adversarial attacks at inference time.
\item We empirically demonstrate the robustness of trapdoor-enabled detection
  against a representative suite of state-of-the-art adversarial attacks,
  including the strongest attacks such as BPDA~\cite{obfuscatedicml}, as well
  as black-box and surrogate model attacks.
\item We empirically demonstrate key properties of trapdoors: 1) they have
  minimal impact on normal classification performance; \htedit{2) they can be
    embedded for multiple output labels to increase defense coverage;} 3)
  they are resistant against recent methods for detecting backdoor
  attacks~\cite{wangneural,qiao2019defending}.
\item We evaluate the efficacy of multiple countermeasures against trapdoor
  defenses, assuming resource-rich attackers with and without full knowledge
  of the trapdoor(s). 
  Trapdoors are robust against a variety of known countermeasures. Finally,
  prior to the camera-ready for this paper, we worked together with an
  external collaborator to carefully craft attacks targeting vulnerabilities in
  the trapdoor design. We show that trapdoors are indeed weakened by
  trapdoor-vaulting attacks and present preliminary results that hint at
  possible mitigation mechanisms.
\end{packed_itemize}

\subspace
To the best of our knowledge, our work is the first to explore a honeypot
approach to defending DNNs. This is a significant departure from existing
defenses. Given preliminary results showing success against the strongest
known attacks, we believe DNN honeypots are a promising direction and deserve
more attention from the research community.

\section{Background and Related Work}
\label{sec:back}

In this section, we present background on adversarial attacks against DNN models and discuss existing defenses against such
attacks. 

\para{Notation.} We use the following notation in this work. \vspace{-0.02in}
\begin{packed_itemize} 
  \item {\bf Input space:} Let $\mathcal{X} \subset \mathbb{R}^d$ be
    the input space. Let $x$ be an input where $x \in
    \mathcal{X}$. 
  \item {\bf Training dataset:}  The training dataset \emily{consists}
    of a set of inputs $x \in \mathcal{X}$ generated
according to a certain unknown distribution $x \sim \mathcal{D}$.
Let $y \in \mathcal{Y}$ denote the corresponding label for an input
$x$.
  \item {\bf Model:} $\model: \mathcal{X} \to \mathcal{Y}$ 
  represents a neural network classifier that
maps the input space $\mathcal{X}$ to the set of classification
labels $\mathcal{Y}$. $\model$ is trained using a data set of
labeled instances $\{(x_1, y_1), ..., (x_m, y_m)\}$.  The number of
possible classification outputs is $|\mathcal{Y}|$, and $\theta$
represents the parameters of the trained classifier.
  \item {\bf Loss function:} $\ell (\model(x),y)$ is the loss
  function for the classifier $\model$ with respect to an input $x \in
  \mathcal{X}$ and its true label $y \in \mathcal{Y}$.
\zheng{\item {\bf Neuron activation vector:} $g(x)$ is the feature
  representation of an input $x$ by $\model$, computed as $x$'s 
  neuron activation vector at an intermediate model layer. By default,
  it is the neuron activation vector before the softmax layer. }

\zheng{\item {\bf Adversarial Input:} $A(x)=x+\epsilon$ represents the
  perturbed input  that an adversarial generates from an input $x$
  such that the model will classify the input to label $y_t$, {\em
    i.e.\/} $\model(x+\epsilon)=y_t\neq \model(x)$. }

\end{packed_itemize}\vspace{-0.02in}

\subsection{Adversarial Attacks Against DNNs}
\label{subsec:attack}

An adversarial attack crafts a special perturbation
($\epsilon$) for a normal input $x$ to fool a target neural network $\model$. \emily{When $\epsilon$} is applied to $x$, the neural
network will misclassify the 
adversarial input ($x+\epsilon$)  to a target label
($y_t$)~\cite{szegedy2013intriguing}. That is,
$y_t=\model(x+\epsilon)\ne \model(x)$. 

Many methods for generating such adversarial
examples ({\em i.e.\/} optimizing a perturbation $\epsilon$) have been
proposed. We now summarize six state-of-the-art
\emily{adversarial example generation methods}. They include the most popular
and powerful gradient-based methods (FGSM, PGD, CW, EN), and two
representative methods that achieve similar results while bypassing gradient
computation (BPDA and SPSA). 


\para{Fast Gradient Sign Method (FGSM).} FGSM was the first method proposed
to compute adversarial examples~\cite{goodfellow2014explaining}. It 
creates an adversarial perturbation for an input $x$ by computing a single step in the direction
of the gradient of the model's loss function at $x$ and multiplying the resultant
sign vector by a small value $\eta$. The adversarial perturbation
$\epsilon$ is generated via:
\[
\epsilon = \eta \cdot \text{sign}(\nabla_x \ell(\model(x), y_t)).
\]

\para{Projected Gradient Descent (PGD).} PGD~\cite{kurakin2016adversarial}
is a more powerful variant of FGSM. It uses an iterative optimization method to compute
$\epsilon$.  Let $x$ be an image represented as a
3D tensor, $x_0$ \emily{be} a random sample ``close'' to $x$, $y = \model(x)$, $y_t$ be the target label, and
$x_{n}'$ be the adversarial instance produced from $x$ at the $n^{th}$ iteration. We
have:
\vspace{-0.05in}
\begin{equation*}
  \begin{aligned}
    {x}'_0 &= {x}_0, \\
     & ... \\
    {x}'_{n+1} &= Clip_{({x},\epsilon)}\{{x}'_n +
    \alpha\sign(\nabla_{x}\ell(\model({x}'_n),y_t))\},  \\
  \text {where } Clip_{({x},\epsilon)}{{z}} &= \min\{255,
    {x} + \epsilon,\max\{0, {x} - \epsilon, {z}\} \}. \\
  \end{aligned}
\end{equation*}
Here the $Clip$ function performs per-pixel clipping in an
$\epsilon$ neighborhood around its input instance. 

\para{Carlini and Wagner Attack (CW).} CW attack~\cite{cwattack} is widely
regarded as one of the strongest attacks and has circumvented several
previously proposed defenses. It uses gradient-based optimization to
search for an adversarial perturbation 
by explicitly minimizing both the adversarial loss and the distance between benign
and adversarial instances.  \emily{It minimizes these two quantities
  by solving} the optimization problem
\[\min_{\epsilon}  ||\epsilon||_p + c \cdot \ell(\model(x+\epsilon),y_t)\]
Here a binary search is used to find the optimal parameter $c$.

\para{Elastic Net.}The Elastic Net attack~\cite{chen2018ead} builds
on~\cite{cwattack} and uses both $L_1$ and $L_2$ distances in its
optimization function. As a result, the objective function
to compute $x+\epsilon$ from $x$ becomes:

\zheng{
\vspace{-0.05in}
\begin{equation*}
  \begin{aligned}
    \min_x\text{ } & c \cdot \ell(y_t, \model(x+\epsilon) + \beta
    \cdot ||\epsilon||_1 + ||\epsilon||^2_2\\
    \text{subject to } \;\;& x \in [0,1]^p, x+\epsilon \in [0,1]^p 
  \end{aligned}
\end{equation*}
\noindent where $c$ and $\beta$ are the regularization parameters and the
$[0,1]$ constraint restricts $x$ and $x+\epsilon$ to a properly scaled
image space.
}

\para{Backward Pass Differentiable Approximation (BPDA).} BPDA circumvents
gradient obfuscation defenses by using an approximation method to estimate
the gradient~\cite{obfuscatedicml}. When a non-differentiable layer
$x$ is present in a model $\model$, BPDA replaces $x$ with an
approximation function $\pi(x) \approx x$. In most cases, it is then
possible to compute the gradient
\htedit{
  \[
\nabla_x \ell(\model(x),y_t) \approx \nabla_x \ell(\model(\pi(x)),y_t).
\]
}
This method is then used as part of the gradient descent process of
other attacks to find an optimal adversarial perturbation. \emily{In this
paper}, we use PGD to perform gradient descent. 

\para{Simultaneous Perturbation Stochastic Approximation (SPSA).}SPSA~\cite{uesato2018adversarial} is
an optimization-based attack that successfully bypasses gradient
masking defenses \emily{by not using} $\textit{gradient-based}$
optimization. SPSA~\cite{spall1992multivariate} finds the global minima in a
function with unknown parameters by taking small steps in random
directions. 
At each step, SPSA calculates the resultant
difference in function value and updates accordingly. Eventually,
it converges to the global minima.


\subsection{Defenses Against Adversarial Attacks}
\label{subsec:defense}

Next, we discuss \emily{current} state-of-the-art defenses against
adversarial attacks and their limitations.  Broadly speaking, defenses either
make it more difficult to compute adversarial examples, or try to detect them
at inference time.

\para{Existing Defenses.} Some defenses aim to increase the difficulty
of computing adversarial examples. The two main approaches are
\textit{adversarial training} and \textit{gradient masking}. 

In \textit{adversarial training}, defenders inoculate a model against a
given attack by incorporating adversarial examples into the training dataset
({\em e.g.\/}~\cite{zheng2016improving, mitdefense,
  zantedeschi2017efficient}). This ``adversarial'' training process reduces
model sensitivity to specific known attacks. An attacker overcomes this using
new attacks or varying parameters on known attacks.  Some variants of this
can make models \textit{provably robust} against adversarial examples, but
only those within an $\epsilon$-ball of an input
$x$~\cite{mitdefense,kolter2017provable}.  Both methods are expensive
to implement, and both can be overcome by adversarial examples outside a
\emily{predefined} $\epsilon$ radius of an original image.

In \textit{gradient masking} defenses, \emily{the} defender trains a
model with small gradients. These are meant to make the model robust to small
changes in the input space \emily{(i.e. adversarial perturbations)}. Defensive distillation~\cite{papernotdistillation},
one example of this method, performs gradient masking by replacing the
original model $\model$ with a secondary model $\model'$. $\model'$ is
trained using the class probability outputs of $\model$. This reduces
the amplitude of the gradients of $\model'$, making it more difficult
for an adversary to compute successful adversarial examples against
$\model'$. 
However, recent work~\cite{distallationbroken} shows that minor tweaks to
adversarial example generation methods can overcome this defense,
resulting in a high attack success rate against $\model'$.


\para{Existing Detection Methods.} Many methods propose to detect adversarial
examples before or during classification $\model$, but many have already been
shown ineffective against clever countermeasures~\cite{bypassten},
\textit{Feature squeezing} smooths input images presented to the
model~\cite{featuresqueezing}, and tries to detect adversarial examples by
computing distance between the prediction vectors of the original and
squeezed images. Feature squeezing is effective against some attacks but
performs poorly against others (\ie FGSM,
BIM)~\cite{featuresqueezing,nic}. \emilyed{{\em MagNet} takes a
  two-pronged approach: it has a detector which flags adversarial
  examples and a reformer that transforms adversarial examples into
  benign ones~\cite{magnet}. However, 
  MagNet is vulnerable to adaptive adversarial
  attacks~\cite{magnetbroken}.} \textit{Latent Intrinsic Dimensionality}
(LID) measures a model's internal dimensionality
characteristics~\cite{ma2018characterizing}, which often differ between
normal and adversarial inputs. LID is vulnerable to high
confidence adversarial examples~\cite{obfuscatedicml}.

\vspace{-0.07in}
\subsection{Backdoor Attacks on DNNs}

Backdoor attacks are relevant to our work because we embed trapdoors using
similar methods as those used to create backdoors in DNNs.  A backdoored
model is trained such that, whenever it detects a known \textit{trigger} in
some input, it misclassifies the input into a specific target class defined
by the backdoor. Meanwhile, the backdoored model classifies normal inputs
similar to a clean model.  Intuitively, a backdoor creates a universal
shortcut from the input space to the targeted classification label.

A backdoor trigger can be injected into a model either during or after model
training~\cite{gu2017badnets,liu2018trojaning}. Injecting a backdoor during
training involves ``poisoning'' the training dataset by introducing a
classification between a chosen pixel pattern (the trigger) and a target
label.  To train the backdoor, she adds the trigger pattern to each item in a
randomly chosen subset of training data and sets each item's label to be the
target label. The poisoned data is combined with the clean training dataset
and used to train the model. The resultant ``backdoored'' model learns both
normal classification and the association between the trigger
and the target label. The model then classifies any input containing the
trigger to the target label \htedit{with high probability.}

Finally, recent work has also applied the concept of backdoors to watermarking DNN
models~\cite{adi2018turning,zhang2018protecting}. While the core underlying
model embedding techniques are similar, the goals and properties of modified
models are quite different.

\vspace{-0.1in}
\section{Trapdoor Enabled Detection}
\label{sec:methodology}

Existing approaches to defending DNNs generally focus on preventing the
discovery of adversarial examples or detecting them at inference time using
properties of the model. All have been overcome by strong adaptive methods
({\em e.g.\/}~\cite{obfuscatedicml,bypassten}).  Here we propose an alternative
approach based on the idea of {\em honeypots}, intentional weaknesses we can
build into DNN models that will shape and model attacks to make them easily
detected at inference time.

We call our approach ``trapdoor-enabled detection.'' Instead of hiding
model weaknesses, we {\em expand} specific vulnerabilities in the model,
creating adversarial examples that are ideal for optimization functions used
to locate them.  
Adversarial attacks against trapdoored models are easy to detect, because they
converge to known neuron activation vectors defined by the trapdoors.

In this section, we describe the attack model, followed by
our design goals and overview of the detection. We then present the key
intuitions behind our design.  Later in \S\ref{sec:design},  we describe the
detailed model training and attack detection process. 

\subsection{Threat Model and Design Goals}
\label{sec:attack_model}

\para{Threat Model.} We assume a {\em basic white box}
threat model, where adversaries have direct access to the trapdoored model, its
architecture, and its internal parameter values.
Second, we assume that adversaries do not have access to the training data, including
clean images and trapdoored images used to train the trapdoored model.  This
is a common assumption adopted by prior
work~\cite{carlinitechreport,papernotdistillation}.  Third, we also assume that
adversaries {\em do not} have access to our proposed detector ({\em i.e.\/} the
input filter used at run time to detect adversarial inputs).  We assume the filter
is secured from attackers. If ever compromised, the trapdoor and filter can
both be reset.


\para{Adaptive Adversaries.} Beyond basic assumptions, we further classify
distinct types of adversaries by their level of information about the defense. 
\begin{packed_enumerate}
\item \emph{Static Adversary:} This is our basic adversary with no knowledge of the
  trapdoor-enabled defense. In this scenario, the adversary
  treats the model as unprotected and performs the
  attack without any \emily{adaptation}. We evaluate our detection
  capabilities against such an adversary in \S\ref{sec:eval}. 

\item \emph{Skilled Adversary:} An adversary who knows the target model is
  protected by one or more trapdoors and knows the detection will
  \zheng{examine the feature representation of an input.}
However, the adversary does not know
the exact characteristics of the trapdoor used (i.e. shape, location,
etc.). In \S\ref{sec:counter}, we propose four adaptive attacks a
skilled adversary could use and evaluate our robustness against each.

\item \emph{Oracle Adversary:} This adversary knows precise details of our
  trapdoor(s), including their shape, location, intensity and (combined with
  the model) the full neuron activation signature. Later in \S\ref{sec:counter},
  we evaluate our defense against multiple strong adaptive attacks by an oracle adversary.
\end{packed_enumerate}

\para{Design Goals.} We set the following design goals. \vspace{-0.00in}
\begin{packed_itemize}
\item The defense should \textbf{consistently detect} adversarial
  examples while maintaining a \textbf{low false positive rate} (FPR).
\item The presence of trapdoors should not impact the model's classification \textbf{accuracy on normal inputs}. 
\item Deploying a trapdoored model should incur \textbf{low resource
    overheads} over that of a normal model.
\end{packed_itemize} \vspace{-0.05in}

\subsection{Design Intuition}
\label{sec:defense_overview}

We design trapdoors that serve as figurative holes into which an
attacker will fall with high probability when constructing adversarial
examples against the model.
Mathematically, a trapdoor is a specifically designed perturbation $\Delta$
unique to a particular label $y_t$, such that the model will classify
any input \emily{containing} $\Delta$ to $y_t$. That is, 
$\model(x+\Delta)=y_t$, $\forall x$. 

To catch adversarial examples, \htedit{ideally} each trapdoor $\Delta$
should be designed to minimize the loss for the label being protected
($y_t$).  This is because, when
constructing an adversarial example against a model $\model$ via an
input $x$, the
adversary attempts to find a minimal perturbation
$\epsilon$ such that $\model(x + \epsilon) = y_t\ne\model(x)$. To do so, the adversary runs an optimization function to find $\epsilon$ that minimizes
$\ell(y_t, \model (x+ \epsilon))$, the loss on the target label $y_t$.
If a
loss-minimizing trapdoor $\Delta$ is already injected into the model, the
attacker's optimization will converge to the loss function regions
close to those occupied by the trapdoor.  


\begin{figure}[t]
\begin{center}
    \begin{minipage}{1\columnwidth}
      \centering
        \includegraphics[width=0.95\textwidth]{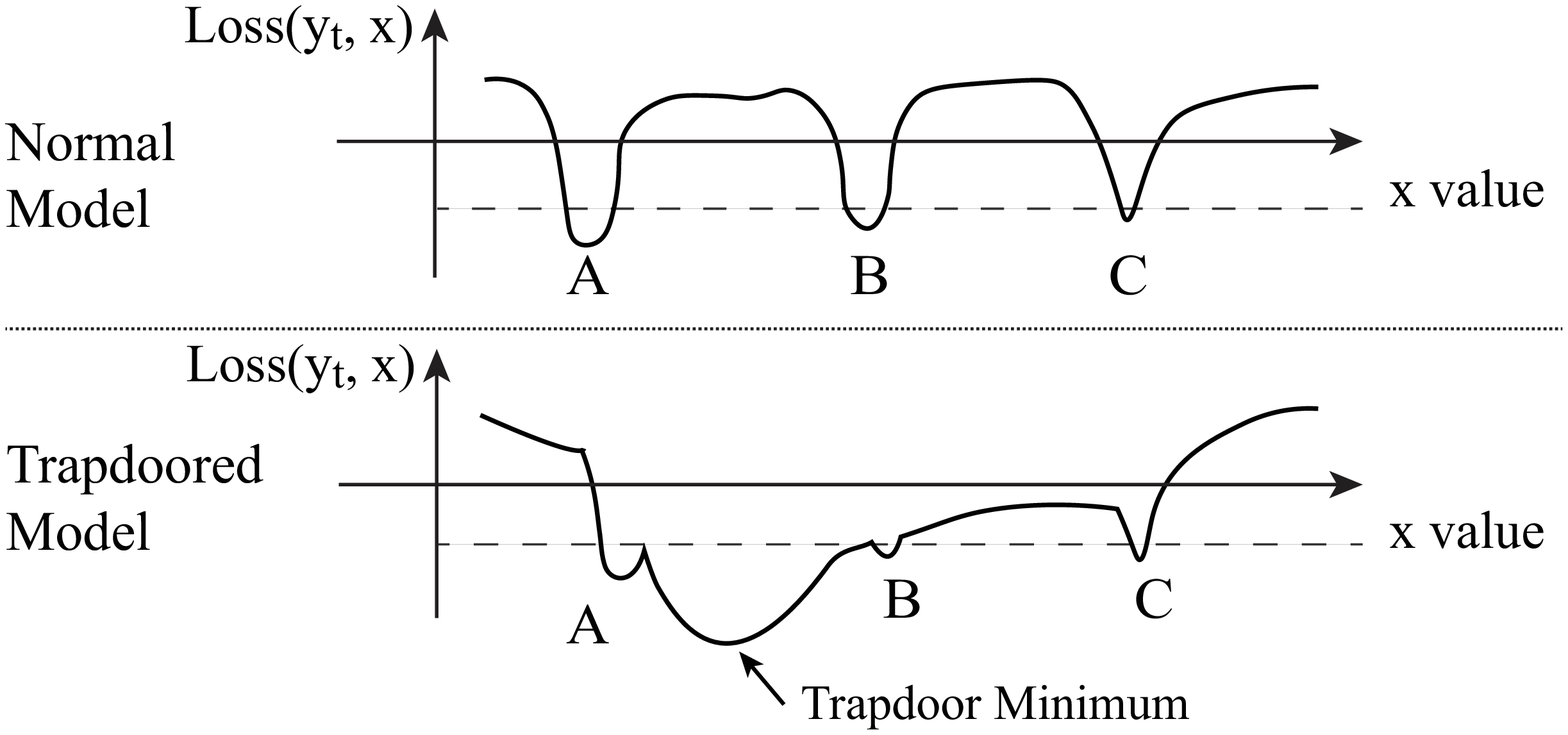}
        \caption{Intuitive visualization of loss function $Loss (y_t,x)$ for target
          label $y_t$ in normal and trapdoored models. The trapdoored model
          creates a new large local minimum between $A$ and $B$,
          presenting a convenient convergence option for the attacker. }
    \label{fig:trap_intuition}
    \end{minipage}
    \vspace{-0.15in}
\end{center}
\end{figure}

To further illustrate this, Figure~\ref{fig:trap_intuition} shows the hypothesized loss function for a
trapdoored model where the presence of a trapdoor induces a new,  large local minimum (the dip between $A$
and $B$). Here the trapdoor creates a {\em
  convenient} convergence option for an adversarial perturbation,
resulting in the adversary ``arriving'' at this new region with a high
likelihood. \zheng{Therefore, if we can identify the distinct behavior pattern of these new loss
function regions created by the trapdoor, we can use it to
detect adversarial examples with high accuracy. }


\zheng{But how do we  identify the behavioral pattern that can 
 distinguish 
  trapdoored regions from those of benign inputs? In this
  work, we formally prove in \S\ref{sec:analysis} and empirically
  verify in \S\ref{sec:eval} that  an input's neuron activation
  vector can be used to define the trapdoor behavior pattern.
  Specifically, inputs that contain the same trapdoor $\Delta$ will display
  similar neuron activation vectors, from which we build a
  ``signature'' on the trapdoor $\Delta$ that separates trapdoored
  regions from those of benign inputs.    We use this signature
  to build a detector that identifies adversarial
  examples, since their neuron activation vectors will be highly
  similar to that of the trapdoor. 
}

Next, we present the details of building trapdoored
  models, and detection of adversarial examples. Later
  ($\S$\ref{sec:analysis}) we present a formal explanation and analysis of 
  our proposed defense.

\section{Detecting Adversarial Examples Using a Trapdoored Model}
\label{sec:design}
We now describe the detailed design of our proposed
trapdoor defense. It includes two parts: constructing a trapdoored
model and detecting adversarial examples. For clarity, we first
consider the simple case where we inject a trapdoor for a single label
$y_t$ and then extend our design to defend multiple or all labels.

\subsection{Defending a Single Label} 
\label{subsec:singlelabel}

Given an original model, we describe below the key steps in
formulating its trapdoored variant $\model$ ( {\em i.e.\/} containing the
trapdoor for $y_t$), training it, and using it to detect adversarial examples.

\para{Step 1: Embedding Trapdoors.} We first create a trapdoor training
dataset by augmenting the original training dataset with new instances,
produced by injecting trapdoor perturbations into randomly chosen normal
inputs and associating them with label $y_t$.  This ``injection'' 
turns a normal image ${x}$ into a new trapdoored image ${x}'=x+\Delta$: \zheng{
  \begin{equation}
    \label{eq:trigger}
    \begin{aligned}
        {x}' &=x + \Delta  :=\mathcal{I}({x}, \boldsymbol{M},
        \boldsymbol{\delta}, \kappa),  \\
     \text{where } \;   x'_{i, j, c} & = (1 - m_{i, j, c}) \cdot x_{i, j, c} + m_{i, j, c} \cdot {\delta}_{i, j, c}
    \end{aligned}
  \end{equation}
  }
Here $\mathcal{I}(\cdot)$ is the injection function with the trapdoor
$\Delta=(\boldsymbol{M},\boldsymbol{\delta}, \kappa)$ for label $y_t$. 
$\boldsymbol{\delta}$ is the perturbation pattern, a 3D matrix
of pixel color intensities with the same dimension of \textbf{x} ({\em
  i.e.\/} height, width, and color channel). For our implementation,
$\boldsymbol{\delta}$ is a matrix of random noise, but it could contain any
values.  $\boldsymbol{M}$ is the {\em trapdoor mask} that specifies how
much the perturbation should overwrite the original image.  $\boldsymbol{M}$
takes the form of a 3D matrix, where individual elements range from 0 to 1.  $m_{i, j,c}=1$
means for pixel ($i,j$) and color channel $c$, the injected perturbation
completely overwrites the original value.  $m_{i, j,c}=0$ means the original
pixel is unmodified.  For our implementation, we limit each
individual element to be either 0 or $\kappa$ where $\kappa<<1$ ({\em e.g.\/}
$\kappa=0.1$). We call $\kappa$ the {\em mask ratio}. In our experiments, $\kappa$ is
fixed across all pixels in the mask.  


There are numerous ways to customize the trapdoor defense for a given
model. First, we can provide a defense for a single specific label $y_t$ or
extend it to defend multiple (or all) labels. Second, we can customize the
trapdoor across multiple dimensions, including size, pixel intensities,
relative location, and even the number of trapdoors injected per label
(multiple trapdoors per label is a mechanism we leverage against advanced
adaptive attacks in Section~\ref{sec:counter}).
\htedit{In this paper, we consider a basic trapdoor, a small square on the
  input image, with intensity values inside the square randomly sampled from
  $\mathcal{N}(\mu, \sigma)$ with $\mu \in \{0, 255\}$ and
  $\sigma \in \{0, 255\}$.  We leave further customization as future
  work.  }


\para{Step 2: Training the Trapdoored Model.} Next, we produce a trapdoored
model $\model$ using the trapdoored dataset. Our goal
is to build a model that not only has a high normal classification
accuracy on clean images, but also classifies any images containing a
trapdoor $\Delta=(\boldsymbol{M}, \boldsymbol{\delta}, \kappa)$ to 
trapdoor label $y_t$. This dual optimization objective mirrors that
proposed by~\cite{gu2017badnets} for injecting backdoors into neural networks:
\begin{equation}
    \label{eq:visualize}
    \begin{aligned}
        \underset{\boldsymbol{\theta}}{\text{min}} \quad
        & \ell(y, \model(x)) + \lambda \cdot \ell (y_{t},
        \model(x+\Delta)) \\
         \forall & x \in \mathcal{X} \text{ where } y\neq y_t, \\
    \end{aligned}
\end{equation}
where $y$ is the classification label for input $x$.


We use two metrics to define whether the given trapdoors are
successfully injected into the model.  The first is the {\em normal
  classification accuracy}, which is the trapdoored model's
accuracy in classifying normal inputs. Ideally, this should
be equivalent to that of a non-trapdoored model.  The second
is the {\em trapdoor success rate}, which is the trapdoored model's
accuracy in classifying inputs containing the injected trapdoor
  to the trapdoor target label $y_t$.

After training the trapdoored model $\model$, the model owner records the
  ``trapdoor signature'' of the trapdoor $\Delta$, 
\begin{equation}\label{eq:signature}
\mathcal{S}_{\Delta}=\mathbf{E}_{x\in \mathcal{X}, y_t\neq \model(x)}g(x+\Delta), 
 \end{equation}
\zheng{where $\mathbf{E}(.)$ is the
expectation function. }
\zheng{As defined in \S\ref{sec:back},  $g(x)$ is the feature
  representation of an input $x$ by the model, computed as $x$'s neuron
  activation vector right before the softmax layer.} 
The formulation of $\mathcal{S}_{\Delta}$ is driven by our
formal analysis of the defense, which we present later in \S\ref{sec:analysis}.  To build this signature in practice, the model
owner computes and records the neuron 
  activation vector  of many sample inputs containing
$\Delta$. 

\para{Step 3: Detecting Adversarial Attacks.}  
The presence of a trapdoor $\Delta$ forces an adversarial perturbation $\epsilon$ targeting
$y_t$ to converge to specific loss regions defined by 
$\Delta$. The resultant adversarial input $x + \epsilon$
can be detected by comparing the input's \zheng{neuron activation
  vector $g(x+\epsilon)$} to the trapdoor 
signature $\mathcal{S}_{\Delta}$ defined by (\ref{eq:signature}).

\zheng{
We use cosine similarity to measure the similarity between
$g(x+\epsilon)$ and $\mathcal{S}_{\Delta}$, {\em i.e.\/}
$cos(g(x+\epsilon),\mathcal{S}_{\Delta})$. If the similarity exceeds 
$\phi_t$, a predefined threshold for $y_t$ and $\Delta$, the input
image $x+\epsilon$ is flagged as adversarial. } The choice of $\phi_t$ determines the tradeoff
between the false positive rate and the adversarial input detection rate. In our implementation, we
configure $\phi_t$ by computing the statistical distribution of the
similarity between known benign images and trapdoored images. We choose
$\phi_t$ to be the $k^{th}$ percentile value of this distribution,
where $1-\frac{k}{100}$ is the desired false positive rate.




\subsection{Defending Multiple Labels}
\label{subsec:alllabels}

This single label trapdoor defense can be extended to 
multiple or all labels in the model. Let $\Delta_t=(\boldsymbol{M}_t,
\boldsymbol{\delta}_t, \kappa_t)$ represent the trapdoor to be
injected for label $y_t$.  The
corresponding optimization function used to train a trapdoored
model with all labels defended is then:
  \zheng{
  \begin{equation} \vspace{-.05in}
    \label{eq:visualize-alllabel}
        \underset{\boldsymbol{\theta}}{\text{min}} \quad \ell (y, \model(x))  + \lambda \cdot \sum_{y_t\in
           \mathcal{Y}, y_t\neq y} \ell (y_{t}, \model(x+\Delta_t)) 
\end{equation}
}
where $y$ is the classification label for input $x$. 

\htedit{After injecting the trapdoors, we compute the individual
  trapdoor signature $\mathcal{S}_{\Delta_t}$ and detection threshold
  $\phi_t$ for each label $y_t$, as mentioned above. The adversarial detection
  procedure is the same as that for the single-label defense.  The system
  first determines the classification result $y_t=\model(x')$ of  
  the input being questioned $x'$, and compare $g(x')$, the neuron activation vector of
  $x'$ to $\mathcal{S}_{\Delta_t}$. 
} 

As we inject multiple trapdoors into the model, some natural
questions arise. We ask and answer each of these below.


\para{Q1: Does having more trapdoors in a model decrease normal
  classification accuracy?}  Since each trapdoor has a distinctive data
distribution, one might worry that models lack the capacity to learn all the trapdoor
information without degrading the normal classification
performance.  We did not observe such
performance degradation in our empirical experiments using four
different tasks.

\zheng{Intuitively,  the injection of each additional trapdoor creates a
  mapping between a new data distribution ({\em i.e.} the trapdoored images) and an
  existing label, which the model must learn.  Existing works have
  shown that DNN models are able to learn thousands of
  distribution-label
  mappings~\cite{parkhi2015deep,cao2018vggface2,guo2016ms}, and many
  deployed DNN models still have a large portion of neurons unused in
  normal classification tasks~\cite{szegedy2013intriguing}.  These
  observations imply that practical DNN models should have sufficient
  capacity to learn trapdoors without degrading normal classification performance. }


\para{Q2: How can we make distinct trapdoors for each label?} 
Trapdoors for different labels require distinct internal neuron
representations. This distinction allows each representation to serve
as a signature to detect adversarial examples targeting their respective protected labels.
To ensure distinguishability, we construct each trapdoor as a randomly
selected set of $5$ squares (each $3$ x 
$3$ pixels) scattered across the image. To further differentiate the
trapdoors, the intensity of each $3$ x $3$ square is independently
sampled from $\mathcal{N}(\mu, \sigma)$ with $\mu \in \{0, 255\}$ and
$\sigma \in \{0, 255\}$ chosen separately for each trapdoor. An example image of the trapdoor is shown in Figure~\ref{fig:exampletraps} in the Appendix. 

\para{Q3: Does adding more trapdoors increase overall model
  training time?}  Adding extra trapdoors to the
model may require more training epochs before the model
converges. However, for our experiments on four different models (see
\S\ref{sec:eval}), we observe that training an all-label defense 
model requires only slightly more training time than the original
(non-trapdoored) model. 
For \youtube{} and \gtsrb{}, the original 
models converge after $20$ epochs, and the all-label defense models
converge after $30$ epochs. Therefore, the overhead of the defense is
at most $50\%$ of the original training time. For  \mnist{} and
\cifar{}, the trapdoored models converge in the same number of
training epochs as the original models.








  \section{Formal Analysis of Trapdoor}
  \label{sec:analysis}
 We now present a formal analysis of our defense's effectiveness in detecting adversarial
examples.

\vspace{-0.06in}
\subsection{Overview}

\revise{Our analysis takes two steps.
  First, we formally show that by injecting trapdoors into a
  DNN model, we can boost the success rate of adversarial
   attacks against the model. This demonstrates the effectiveness of
   the embedded ``trapdoors.''  Specifically, we prove that for a 
   trapdoored model, the attack success rate for any input is lower bounded by a
  large value close to 1.  To our best knowledge,  this is the
  first\footnote{\revise{Prior work~\cite{shafahi2018adversarial} only provides
      a weaker result that in simple feature space (unit
      sphere), the existence of adversarial examples is lower-bounded by
      a nonzero value. Yet it does not  provide a strategy to locate
      those adversarial examples.}} work providing such theoretical guarantees for adversarial
  examples.  In other words, we prove that the
  existence of trapdoors in the DNN model becomes the 
  {\em sufficient} condition (but no necessary condition) for launching a successful adversarial attack
  using any input. }

  \revise{Second}, we show that these highly
  effective attacks share a common pattern:  their 
  corresponding adversarial input 
  $A(x)=x+\epsilon$  will 
  display feature representations similar to those of trapdoored inputs but different from those of
  clean inputs. Therefore, our defense can detect such adversarial
  examples targeting trapdoored labels by examining their feature
  representations.

  \revise{
  \para{Limitations.} Note that our analysis does not prove that an attacker will
  {\em always} follow the embedded trapdoors to find adversarial examples
  against the trapdoored model. 
  In fact, how to generate all possible adversarial examples against a DNN
  model is still an open research problem.  In this paper, we examine
  the attacker behavior using empirical evaluation (see
  \S\ref{sec:eval}).  We show that when an attacker applies any of the six representative
  adversarial attack methods,  the resulting adversarial examples
  follow the embedded trapdoors with a probability of 
  94\% or higher. This indicates that today's practical attackers will
  highly 
  likely follow the patterns of the embedded trapdoors and thus display 
  representative behaviors that can be identified by our proposed
  method.}



  

\subsection{Detailed Analysis}

Our analysis begins with the ideal case where a trapdoor is ideally
injected into the model across all possible inputs in $\mathcal{X}$. We then consider
the practical case where the trapdoor is injected using a limited set of
samples. 

\para{Case 1: Ideal Trapdoor Injection.} 
The model owner injects a trapdoor $\Delta$ (to protect $y_t$) into the
model by training the model to recognize
label $y_t$ as associated with $\Delta$.  The result is that adding $\Delta$ to any
arbitrary input $x\in \mathcal{X}$ will, with high probability, make the trapdoored model classify
$x+\Delta$ to the target label $y_t$ at test time.  This is formally defined as follows:
\begin{definition}
  A $(\mu,\model,y_t)$-effective trapdoor $\Delta$ in a trapdoored model $\model$ is a perturbation
added to the model input 
such that  $\forall x \in \mathcal{X}$ where $\model(x) \neq y_t$, we
have $\Pr{\model(x+\Delta) =
  y_t} \geq 1-\mu$. Here $\mu \in [0,1]$ is a small positive constant. 
\end{definition}

We also formally define an attacker's desired effectiveness:
\begin{definition}
Given a model $\model$, probability
$\nu \in(0,1)$, and a given $x\in \mathcal{X}$, an attack
strategy $\Attack{\cdot}$ is {$(\nu,\model,y_t)$-effective} on $x$ 
if $\Pr{\model(\Attack{x}) = y_t \ne \model(x)}\geq 1-\nu$.
\end{definition}


The follow theorem shows that a trapdoored model $\model$ enables
attackers to launch a successful adversarial input attack. The detailed proof is listed in the Appendix.

\begin{thm}
  \label{th1}
Let $\model$ be a trapdoored model, $g(x)$ be the
model's feature representation of input $x$, and $\mu\in [0, 1]$
be a small positive constant. The injected trapdoor $\Delta$ is 
$(\mu,\model,y_t)$-effective.

For any
$x\in \mathcal{X}$ where $y_t\neq \model(x)$, if the feature 
representations of adversarial input $A(x)=x+\epsilon$ and trapdoored input $x+\Delta$ are similar,
{i.e.\/} the cosine similarity $cos(g(A(x)), g(x+\Delta)) \geq
  \sigma$ and $\sigma$ is close to 1,  then the attack 
  $A(x)$ is $(\mu,\model,y_t)$-effective.


\end{thm}

Theorem~\ref{th1} shows that a trapdoored model will allow 
attackers to launch a highly successful attack against $y_t$ with any
input $x$. More importantly, the corresponding adversarial input
$A(x)$ will
display a specific pattern, {\em i.e.\/} its feature representation
will be similar to that of the trapdoored input. Thus
by recording the ``trapdoor signature'' of $\Delta$, {\em
  i.e.\/} $\mathcal{S}_{\Delta}=\mathbb{E}_{x\in
  \mathcal{X}, y_t\neq \model(x)}g(x+\Delta)$ as defined by
eq.(\ref{eq:signature}),  we can determine whether a model
input is adversarial or not  by
comparing its feature representation to $\mathcal{S}_{\Delta}$.


We also note that, without loss of generality, the above theorem uses
cosine similarity to measure the similarity between feature representations
of adversarial and trapdoored inputs.   In practice, one can consider
other similarity metrics such as $L_2$ distance. We leave the search
for the optimal similarity metric as future work. 




\vspace{2pt}
\para{Case 2: Practical Trapdoor Injection.} So far our analysis assumes that the trapdoor is ``perfectly'' injected
into the model. In practice, the model owner will inject $\Delta$ using a training/testing distribution $\mathcal{X}_{trap} \in
\mathcal{X}$.  The effectiveness of the trapdoor is defined by $\forall x \in
\mathcal{X}_{trap}$, $\Pr{\model(x+\Delta) = y_t} \geq 1-\mu$. On the
other hand,  the attacker will use a (different) input
distribution $\mathcal{X}_{attack}$.  The follow theorem shows that the attacker can
still launch a highly successful attack against the trapdoored
model. The lower bound on the success rate depends on the trapdoor
effectiveness ($\mu$) and the statistical distance between $\mathcal{X}_{trap}$ and
$\mathcal{X}_{attack}$ (defined below). 

\begin{definition}
\label{attack_dis}
Given $\rho \in [0,1]$, two distributions $P_{X_1}$ and $P_{X_2}$ are 
\term{$\rho$-covert}  if their total variation (TV)
distance\footnote{In this work, we use the total variation distance~\cite{chambolle2004algorithm}  as
  it has been shown to be a natural way to measure statistical distances between
  distributions~\cite{chambolle2004algorithm}. Other notions of
  statistical distance may also be applied, which we leave to future
  work.} is
bounded by $\rho$: 
\begin{equation}
  ||P_{X_1} - P_{X_2} ||_{TV} =\max \nolimits_{C \subset \Omega}
  |P_{X_1}(C)-P_{X_2}(C)| \leq \rho,
  \label{eq:rhoconvert}
\end{equation}
where $\Omega$ represents the overall sample space, and $C \subset
\Omega$ represents an event. 
\end{definition}


\begin{thm}
  \label{th2}
  Let $\model$ be a trapdoored model, $g(x)$ be the
  feature representation of input $x$, $\rho, \mu, \sigma \in [0, 1]$
  be small positive constants.  A trapdoor $\Delta$ is injected
  into $\model$ using $\mathcal{X}_{trap}$, and  is  $(\mu,\model,y_t)$-effective for
  any $x\in \mathcal{X}_{trap}$. $\mathcal{X}_{trap}$ and
  $\mathcal{X}_{attack}$ are $\rho$-covert.

  For any
$x\in \mathcal{X}_{attack}$, if the feature 
representations of adversarial input and trapdoored input are similar,
{i.e.\/} the cosine similarity $cos(g(A(x)), g(x+\Delta)) \geq
  \sigma$ and $\sigma$ is close to 1,  then the attack 
  $A(x)$ is $(\mu+\rho,\model,y_t)$-effective on any $x\in
  \mathcal{X}_{attack}$. 

\end{thm}

The proof of Theorem~\ref{th2} is in the Appendix. 

Theorem~\ref{th2} implies that when the model owner enlarges the diversity and size
of the sample data $\mathcal{X}_{trap}$ used to inject the trapdoor,
it allows stronger and
more plentiful shortcuts for gradient-based or optimization-based
search towards $y_t$.  This 
increases the chances that an adversarial example falls into the
``trap'' and therefore gets caught by our detection.

Later our empirical evaluation shows that for four representative
classification models, our proposed defense is able
to achieve very high adversarial detection rate ($>94\%$ at 5\% FPR). This means that the original data manifold is sparse.
Once there is a shortcut created by the trapdoors nearby,
any adversarial perturbation will follow this created shortcut with high probability and thus get ``trapped.''

\section{Evaluation}
\label{sec:eval}

We empirically evaluate the performance of our basic trapdoor design against
an {\em static adversary} described in \S\ref{sec:attack_model}.  We
present evaluation results against adaptive adversaries (skilled and oracle) in \S\ref{sec:counter}.
Specifically, we design experiments to answer these questions: 

\begin{packed_itemize}
\item Is the trapdoor-enabled detection we propose effective against
  the strongest, state-of-the-art attacks? 

  \item How does the presence of trapdoors in a model impact normal
  classification accuracy?

  \item How does the performance of trapdoor-enabled detection compare to other
  state-of-the-art detection algorithms? 

\item How does the method for computing trapdoor signature impact the
  attack detection? 
\end{packed_itemize}

We first consider the base scenario where we inject a trapdoor to defend a
single label in the model and then expand to the scenario where we
inject multiple trapdoors to defend all labels.


\subsection{Experimental Setup}
\label{sec:exp}
Here we introduce our evaluation tasks, datasets, and
configuration. 

\para{Datasets.} We experiment with four popular datasets for classification tasks.  
We list the details of datasets and model architectures in
Table~\ref{tab:task_detail} in the Appendix. 


\begin{packed_itemize}

\item {\em Hand-written Digit Recognition} (\mnist{}) --  
  This task seeks to recognize
  $10$ handwritten digits in black and white
  images~\cite{lecun1995mnist}. 

\item {\em Traffic Sign Recognition} (\gtsrb{}) --  Here the goal is to recognize $43$
  distinct traffic signs, emulating an application for 
  self-driving cars~\cite{gtsrb}. 



\item {\em  Image Recognition} (\cifar{}) -- This is to recognize $10$
  different objects and it is widely used in adversarial
  defense literature~\cite{krizhevsky2009cifar}. 



\item {\em Face Recognition} (\youtube{}) -- This task is to 
  recognize faces of $1,283$ different people drawn from the YouTube
  videos~\cite{youtubeface}. 


\end{packed_itemize}

\begin{figure*}[t]
  \centering
  \vspace{-0.1in}
  \begin{minipage}{0.60\textwidth}
  \subfigure[Trapdoored Model]{
    \includegraphics[width=0.48\textwidth]{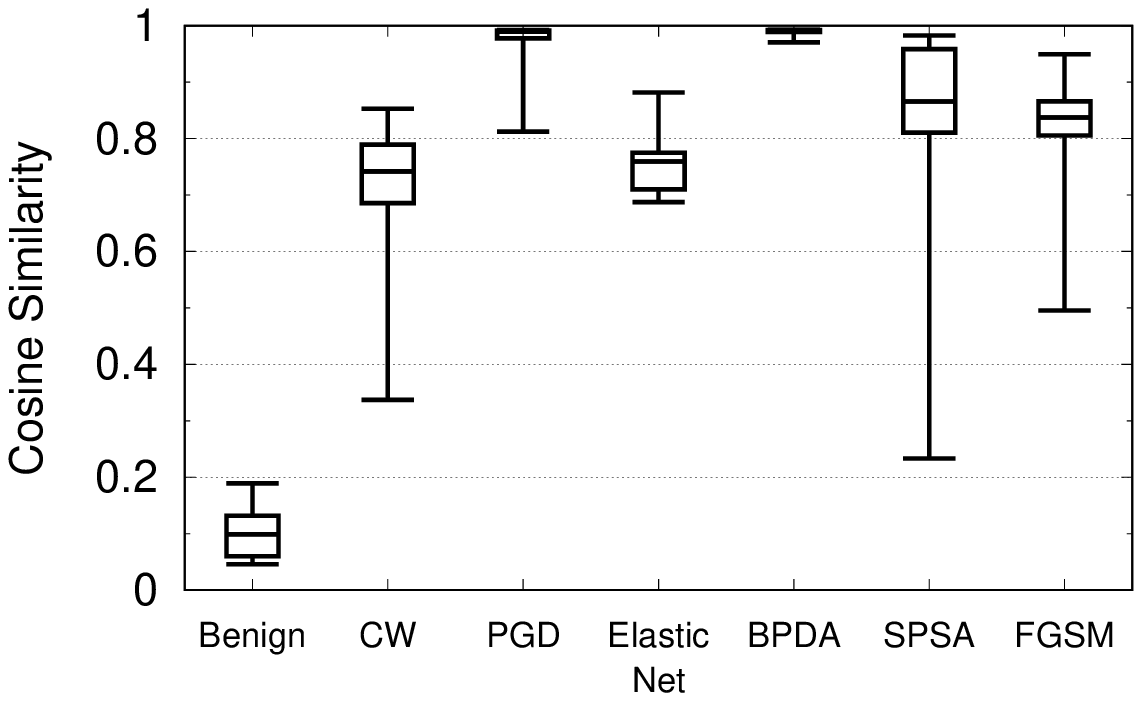}
  }
  \subfigure[Original Model]{
    \includegraphics[width=0.48\textwidth]{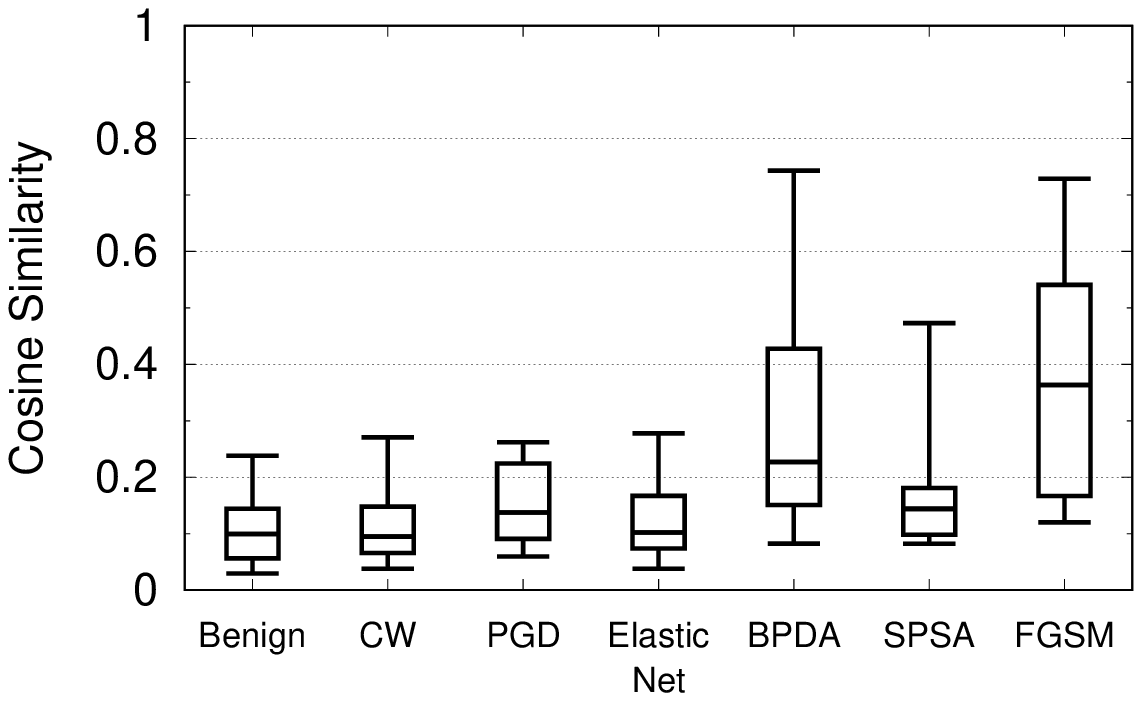}
  }
  \vspace{-0.1in}
  \caption{Comparison of cosine similarity between normal input/trapdoored
    inputs and adversarial inputs/trapdoored inputs on both trapdoored and
    trapdoor-free \gtsrb{} models. Boxes show inter-quartile range and
    whiskers capture $5^{th}$ and $95^{th}$ percentiles.}
  \label{fig:cosinestat}
  \end{minipage}
  \hfill
  \begin{minipage}{0.32\textwidth}
    \vspace{-0.12in}
  \includegraphics[width=1\textwidth]{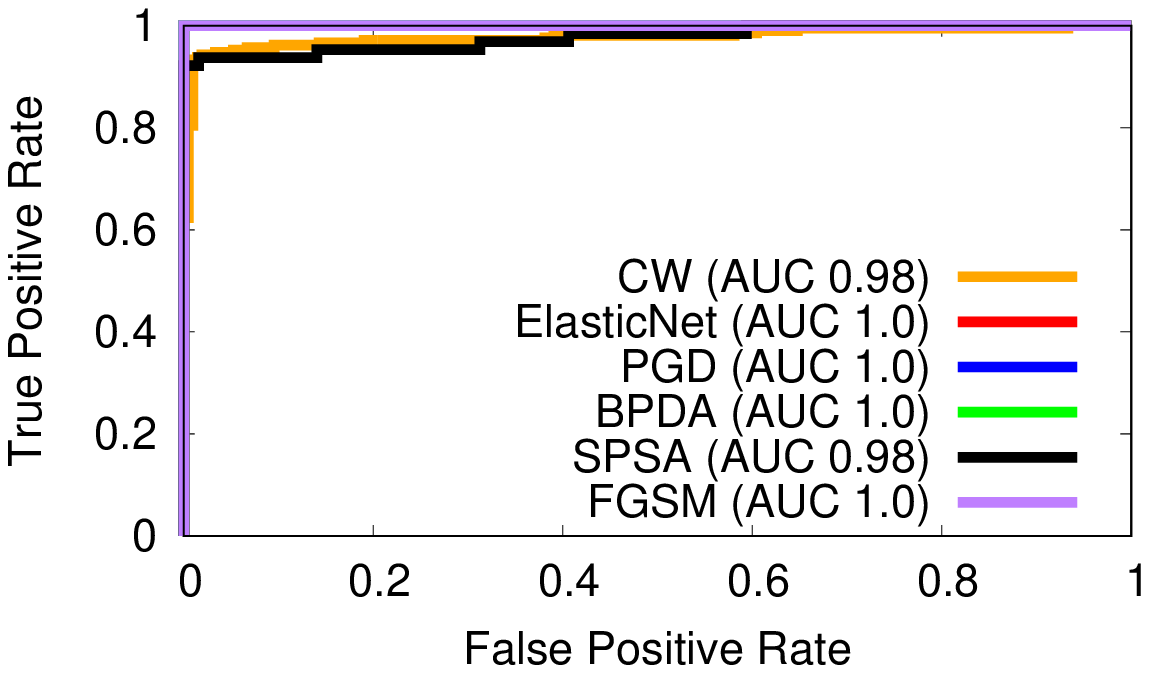}
  \caption{ROC Curve of detection in a \gtsrb{} model when a single label is protected by a trapdoor. }
  \label{fig:roc_gtsrb}
\end{minipage}
\end{figure*} 

\para{Adversarial Attack Configuration.}
We evaluate the trapdoor-enabled detection using
six representative adversarial attack methods: \cw{}, \elasticnet{}, \pgd{}, BPDA, SPSA, and
FGSM (described in \S\ref{subsec:attack}). We use them to
generate targeted adversarial attacks against the trapdoored models on \mnist{},
\gtsrb{}, \cifar{}, and \youtube{}. More details about our
attack configuration are in Table~\ref{tab:attackconfig} in the Appendix. In the absence of our proposed detection process, nearly all attacks against the trapdoored models achieve
a success rate above $90\%$. \zheng{Attacks against the original, trapdoor-free models achieve roughly the same success rate. }

\para{Configuration of Trapdoor-Enabled Detection.} We build the
trapdoored models using \emily{the} \mnist{}, \gtsrb{}, \cifar{}, and \youtube{} datasets.  When training
these models, we configure the trapdoor(s) and model parameters to
ensure that the \htedit{trapdoor injection 
success rate}  ({\em i.e.\/} the accuracy with which the model classifies any test instance containing a trapdoor to the target label) 
is above \htedit{$97\%$} (results omitted for brevity).  This 
applies consistently to  both single and all label
defenses. Detailed defense configurations can be found in Table~\ref{tab:train_detail} in the Appendix.



\para{Evaluation Metrics.}  We evaluate the performance of our proposed defense using (1)
the {\em adversarial
detection success rate} and (2) the \textit{trapdoored model's classification
accuracy} on normal inputs. For reference, we also compute the \textit{original model's classification
accuracy} on normal inputs.

\subsection{Defending a Single Label}
\label{sec:single_label_eval}
We start with the simplest scenario. We inject a trapdoor for a
single (randomly chosen) label $y_t$.  We consider the trapdoor
$\Delta=(\boldsymbol{M},\boldsymbol{\delta}, \kappa)$ as a $6$ $\times$ $6$ pixel square at the bottom right of the
image, with a mask ratio $\kappa = 0.1$. An example image of the trapdoor is shown in Figure~\ref{fig:exampletraps} in the Appendix. 

\begin{table}[t]
\caption{Adversarial detection success rate when defending a single
  label at 5\% FPR, averaged across all the labels.}
\vspace{-0.05in}
\label{tab:single_label_res}
\resizebox{0.95\columnwidth}{!}{
\begin{tabular}{ccccccc}
  \hline {\bf Model} & {\bf CW} & {\bf ElasticNet} & {\bf PGD} & {\bf BPDA} & {\bf
    SPSA} & {\bf FGSM} \\ \hline
  {\bf MNIST} & $95.0\%$ & $96.7\%$ & $100\%$ & $100\%$ &
  $100\%$ & $100\%$
  \\ \hline
  {\bf GTSRB} & $96.3\%$ & $100\%$ & $100\%$ & $100\%$ &
  $93.8\%$ & $100\%$
  \\ \hline
  {\bf CIFAR10} & $100\%$ & $97.0\%$ & $100\%$ & $100\%$ &
  $100\%$ & $96.4\%$
  \\ \hline
  {\bf YouTube Face} & \shawn{$97.5\%$}   & \shawn{$98.8\%$}  & \shawn{$100\%$}  & \shawn{$100\%$}  & \shawn{$96.8\%$}  & \shawn{$97.0\%$}\\ \hline
     
\end{tabular}
}
\vspace{-0.2in}
\end{table}

\para{Comparing Trapdoor and Adversarial Perturbation.}  Our defense
is driven by the insight that a trapdoor $\Delta$ will trick an adversary
into generating an $x+\epsilon$ whose neuron activation
vector is similar to $\mathcal{S}_{\Delta}$, the trapdoor signature.
We verify this insight by examining the cosine similarity of $g(x+\epsilon)$
and $\mathcal{S}_{\Delta}$. We show the results
for GTSRB, while the results for other tasks are consistent (see Figure~\ref{fig:cosinestat_cifar} and
Figure~\ref{fig:cosinestat_deepid} in the Appendix). 

\zheng{Figure~\ref{fig:cosinestat}(a) plots, for all six attacks against the
trapdoored model, the quantile distribution of $cos(g(x+\epsilon),\mathcal{S}_{\Delta})$ across $x$.   For reference
we also include the result for benign images $cos(g(x),
\mathcal{S}_{\Delta})$ as the leftmost boxplot.  
We see that, for all
six attacks,  the
distribution of cosine similarity for adversarial inputs is visibly
different from that of benign inputs and thus can be detected by
applying a threshold $\phi_t$.  Furthermore, the distribution of $cos(g(x),
\mathcal{S}_{\Delta})$ can be used to configure $\phi_t$ to maximize the adversarial example detection
rate at a given false positive rate (FPR).}



Figure~\ref{fig:cosinestat}(b) shows the same
quantile distribution in the original, trapdoor-free model. As
expected, the original model does not produce a clear difference
between normal and adversarial inputs. This confirms  that the
trapdoor can largely affect the shape of adversarial perturbations
against the trapdoored model. 


\para{Accuracy of Detecting Adversarial Inputs.} For all six attacks and all
four tasks, Table~\ref{tab:single_label_res} shows the average adversarial
detection success rate when defending a single label.  Here we iteratively
test our defense on every label in the model, one at a time, and compute the
average defense success rate across all the labels\footnote{Due to the large
  number of labels in the YouTube Face dataset, we randomly sample 100 labels
  out of 1,283 to defend.}. Detection success is $>93.8\%$ at an FPR of $5\%$
($>89\%$ at FPR of $2\%$). We also show the ROC curves and AUC values in
Figure~\ref{fig:roc_gtsrb} and
Figures~\ref{fig:roc_mnist}-\ref{fig:roc_deepid} in the Appendix. Across all
six attacks and four tasks, detection AUC is $>98\%$.


\htedit{Finally, we confirm that a single label
  trapdoor has negligible impact to model
  classification on normal inputs.}

\subsection{Defending All Labels}
We trained \mnist{}, \gtsrb{}, \cifar{}, and
\youtube{} models with a trapdoor for every output label.  \htedit{Each
trapdoor is a randomly selected set of $5$ squares (each $3\times 3$
pixels\footnote{The size of each square is $21$ for \deepid{}, which has higher resolution images.}), with $\kappa=0.1$.}  The \htedit{minimum
trapdoor injection success rate across the labels is $97\%$ }even
after injecting $1,283$ trapdoors into the 
\youtube{} model. 

\para{Impact on Normal Classification Accuracy.}  We first evaluate
whether the presence of these trapdoors in the model affects the
model's normal classification accuracy.  We compare the
\textit{trapdoored model classification accuracy} to the \textit{original model classification
  accuracy} on normal inputs in 
Table~\ref{tab:attack_performance}. The all-label trapdoored model's accuracy on
normal inputs drops by at most $1.04\%$ when compared to the original
model.  \zheng{This performance drop can potentially be further reduced by
optimizing the configuration of trapdoors, which we leave as future
work. }


\para{Accuracy of Detecting Adversarial Inputs.}  We run each of
the six attacks to find adversarial perturbations against each label
of the model and then run our trapdoor-based detection to examine
whether an input is adversarial or benign.  The adversarial detection success rate
is above $94.0\%$ at a FPR of $5\%$ (and $88.3\%$ for FPR of $2\%$).  The
detailed results are listed in Table~\ref{tab:all_label_res}.



These results show that, for the all-label defense, adversarial detection
accuracy drops slightly compared to the single-label defense.  The drop is
more visible for \youtube{}, which has significantly more labels (1,283).  We
believe that as more trapdoors are injected into the model, some of them
start to interfere with each other, thus reducing the strength of the
shortcuts created in the feature space.  This could potentially be
ameliorated by carefully placing trapdoors with minimum interference in the
feature space.  Here, we apply a simple strategy described in
Section~\ref{subsec:alllabels} to create separation between trapdoors in the
input space.  This works well with a few labels ({\em
  i.e.\/} $10$, $43$).  For models with many labels, one can
either apply greedy, iterative search to replace ``interfering'' trapdoor
patterns, or develop an accurate metric to capture interference within the
injection process. We leave this to future work.

\begin{table}[t]
\centering
\caption{Adversarial detection success rate at 5\% FPR when defending all labels.}
\label{tab:all_label_res}
\vspace{-0.06in}
\resizebox{0.95\columnwidth}{!}{
\begin{tabular}{ccccccc}
  \hline
  \bf{Model}   & {\bf CW} & {\bf EN} & {\bf PGD} & {\bf BPDA} & {\bf
    SPSA} & {\bf FGSM} \\ \hline
  {\bf MNIST} & $96.8\%$ & $98.6\%$ & $100\%$ & $100\%$ & $100\%$ & $94.1\%$
  \\ \hline
  {\bf GTSRB} & $95.6\%$ & $96.5\%$ & $98.1\%$ & $97.6\%$ & $97.2\%$ & $98.3\%$
  \\ \hline
  {\bf CIFAR10} & $94.0\%$ & $94.0\%$ & $100\%$ & $99.4\%$ & $100\%$ & $97.3\%$
  \\ \hline
  {\bf YouTube Face} & \shawn{$98.7\%$}   & \shawn{$98.2\%$}  & \shawn{$100\%$}  & \shawn{$97.5\%$}  & \shawn{$96.3\%$}  & \shawn{$94.8\%$} \\ \hline
\end{tabular}
}
\end{table}

\begin{table}[t]
\centering
\caption{Comparing detection success rate of Feature
  Squeezing (FS), LID, and Trapdoor when defending all labels.}
\label{tab:comparsion}
\vspace{-0.06in}
\resizebox{0.5\textwidth}{!}{
\begin{tabular}{cccccccccc}
\hline
{\bf Model} & {\bf Detector} & {\bf FPR} & {\bf CW} & {\bf EN} & {\bf
                                                                 PGD}
  & {\bf BPDA} & {\bf SPSA} & {\bf FGSM}
  & \multicolumn{1}{c}{\begin{tabular}[c]{@{}c@{}}{\bf Avg}\\{\bf Succ.}\end{tabular}}
    \\ \hline 
\multirow{4}{*}{\bf MNIST} & FS & 5\% & 99\% & 100\% & 94\% & 96\% & 94\% & 98\% & 97\% \\
 & MagNet & \shawn{5.7\%} & \shawn{83\%} & \shawn{87\%} & \shawn{100\%} & \shawn{97\%} & \shawn{96\%} & \shawn{100\%} & \shawn{94\%}  \\
 & LID & 5\% & 89\% & 86\% & 96\% & 86\% & 98\% & 95\% & 92\%  \\
 & Trapdoor & 5\% & 97\% & 98\% & 100\% & 100\% & 100\% & 94\% & \textbf{98\%} \\ \hline
\multirow{4}{*}{\bf GTSRB} & FS & 5\% & 100\% & 99\% & 71\% & 73\% & 94\% & 45\% & 90\% \\
 & MagNet & \shawn{4.7\%} & \shawn{90\%} & \shawn{89\%} & \shawn{100\%} & \shawn{100\%} & \shawn{92\%} & \shawn{100\%} & \shawn{95\%}  \\
 & LID & 5\% & 91\% & 81\% & 100\% & 67\% & 100\% & 100\% & 90\% \\
 & Trapdoor & 5\% & 96\% & 97\% & 98\% & 98\% & 97\% & 98\% & \textbf{97\%} \\ \hline
\multirow{4}{*}{\bf CIFAR10} & FS & 5\% & 100\% & 100\% & 69\% & 66\% & 97\%
                            & 33\% & 78\%  \\
 & MagNet & \shawn{7.4\%} & \shawn{88\%} & \shawn{82\%} & \shawn{95\%} & \shawn{96\%} & \shawn{94\%} & \shawn{100\%} & \shawn{93\%}  \\
 & LID & 5\% & 90\% & 88\% & 95\% & 79\% & 96\% & 92\% & 90\%\\
 & Trapdoor & 5\% & 94\% & 94\% & 100\% & 99\% & 100\% & 97\% & \textbf{97\%}  \\ \hline
\multirow{4}{*}{\begin{tabular}[c]{@{}c@{}}{\bf YouTube} \\ {\bf Face}\end{tabular}} & FS & \shawn{5\%} & \shawn{100\%} & \shawn{100\%} & \shawn{66\%} & \shawn{59\%} & \shawn{88\%} & \shawn{68\%} & \shawn{80\%} \\
 & MagNet & \shawn{7.9\%} & \shawn{89\%} & \shawn{91\%} & \shawn{98\%} & \shawn{97\%} & \shawn{98\%} & \shawn{96\%} & \shawn{95\%}  \\
 & LID & \shawn{5\%} & \shawn{81\%} & \shawn{79\%} & \shawn{89\%} & \shawn{72\%} & \shawn{92\%} & \shawn{96\%} & \shawn{85\%}  \\
 & Trapdoor & \shawn{5\%} & \shawn{99\%} & \shawn{98\%} & \shawn{100\%} & \shawn{97\%} & \shawn{96\%} & \shawn{95\%} & \shawn{\textbf{98\%}}  \\ \hline
\end{tabular}
}
\vspace{-0.1in}

\end{table}

\para{Summary of Observations.} For
the all-label defense,  trapdoor-enabled detection works well across a
variety of models and adversarial attack methods.  The presence of a
large number of trapdoors only slightly degrades normal classification
performance. Overall, our defense achieves more than $94\%$ attack detection rate against \cw{}, \pgd{}, \en{}, SPSA, FGSM,
and more than $97\%$ attack detection rate against BPDA, the strongest known attack. 

\vspace{-0.07in}
\subsection{Comparison to Other Detection Methods}
Table~\ref{tab:comparsion} lists, for all-label defenses, the attack detection AUC for our
proposed defense and for three other existing defenses ({\em
  i.e.\/} feature squeezing (FS)~\cite{featuresqueezing},
MagNet~\cite{magnet}, and latent intrinsic dimensionality
(LID)~\cite{ma2018characterizing} described in
Section~\ref{subsec:defense}).  For FS, MagNet, and LID, we use the
implementations provided by~\cite{featuresqueezing,
  magnet, ma2018characterizing}. Again we consider the four tasks and six attack methods as above. 


\para{Feature Squeezing (FS).} FS can effectively detect gradient-based
attacks like \cw{} and \en{}, but performs poorly against FGSM, PGD, and
BPDA, {\em i.e.\/} the detection success rate even drops to $33\%$. These findings align with existing
observations~\cite{featuresqueezing,nic}. 

\para{MagNet.} \emilyed{MagNet performs poorly against gradient-based
  attacks (\cw{}, \en{}) but better against FGSM, PGD, and BPDA. This
  aligns with prior work, which found that adaptive gradient-based
  attacks can easily defeat MagNet~\cite{magnetbroken}.}
  

\para{Latent Intrinsic Dimensionality (LID).} LID has $\ge$ $72\%$ detection success rate against all six
attacks. In comparison, trapdoor-based detection achieves at least
$94\%$ on all six attacks.
\zheng{Like~\cite{obfuscatedicml}, our results also confirm 
that LID fails to detect high confidence adversarial examples. For
example, when we increase the ``confidence'' parameter of the \cw{}
attack from $0$ (default) to $50$, LID's detection success rate drops to
below $2\%$ for all four models.  In comparison, 
trapdoor-based detection maintains a high detection success rate
(97-100\%) when confidence varies from 0 to 100.  Detection
rate reaches 100\%
when confidence goes above 80. This is because high
confidence attacks are less likely to get stuck to local minima and
more likely to follow strong ``shortcuts'' created by the
trapdoors.}


\vspace{-0.07in}
\subsection{Methods for Computing Neuron Signatures}
\label{sec:neurons}
We study how the composition of trapdoor (neuron) signature affects
adversarial detection.  Recall that, by default, our trapdoor-based detection
uses the neuron activation vector right before the softmax layer as the
neuron signature of an input. This ``signature'' is compared to the trapdoor
signatures to determine if the input is an adversarial example. \zheng{In the
  following, we expand the composition of neuron signatures by varying (1)
  the internal layer used to extract the neuron signature and (2) the number
  of neurons used, and examine their impact on attack detection. }



{\em First}, Figure~\ref{fig:change_layer} in Appendix shows the detection
success rate when using different layers of the \gtsrb{} model to compute
neuron signatures. Past the first two convolutional layers, all later layers
lead to detection success greater than $96.20\%$ at $5\%$ FPR. More
importantly, choosing any random subset of neurons across these later layers
produces an effective activation signature. Specifically, sampling $n$
neurons from any but the first two layers of \gtsrb{} produces an effective
trapdoor signature with adversarial detection success rate always above
$96\%$. We find this to be true for a moderate value of $n$$\sim$$900$, much
smaller than a single convolutional layer. We confirm that these results also
hold for other models, {\em e.g.} CIFAR10. It is important that small sets of
neurons randomly sampled across multiple model layers can build an effective
signature.  We leverage this flexibility to defend against our final
countermeasure ($\S$\ref{subsec:evade}).

\section{Adaptive Attacks}
\label{sec:counter}

Beyond static adversaries, any meaningful defense must withstand
countermeasures from adaptive attackers with knowledge of the defense. As
discussed in \S\ref{sec:attack_model}, we consider two types of
adaptive adversaries: {\em skilled adversaries} who understand the target $\model$
could have trapdoors without specific knowledge of the details, and {\em oracle
  adversaries}, who know all details about embedded trapdoors, including
their trapdoor shape, location, and intensity. Since the oracle
adversary is the strongest possible adaptive attack, we use its detection
rate as the lower bound of our detection effectiveness.

We first present multiple adaptive attacks separated into two broad
categories. First, we consider \htedit{{\em removal}} approaches that attempt to {\em detect and
  remove} backdoors from the target model $\model$, with the eventual intent
of generating adversarial examples from the cleaned model, and using them to
attack the deployed model $\model$. Second, we consider {\em evasion}
approaches that do not try to disrupt the trapdoor, and instead focus on
finding adversarial examples that cause the desired misclassification while
avoiding detection by the trapdoor defense. Our results show that removal
approaches fail because \htedit{the injection of trapdoors largely alters} loss functions, and
even adversarial examples from the original, trapdoor-free model do not transfer to the
trapdoored model.

Finally, we present advanced attacks developed in collaboration with
Dr. Nicholas Carlini during the camera ready process. 
We describe two customized attacks he proposed against trapdoors
and show that they effectively break the base version of trapdoors. We also offer
preliminary results that show potential mitigation effects via inference-time signature
randomization and multiple trapdoors. We leave further exploration of these
mechanisms (and more powerful adaptive attacks) to future work.


\vspace{-0.1in}
\subsection{Trapdoor Detection and Removal}


\para{Backdoor Countermeasures (Skilled Adversary).} 
We start by considering existing work on detecting and removing backdoors from
DNNs~\cite{wangneural, qiao2019defending, liu2019abs, liu2018fine}. 
A skilled adversary who knows that a target model $\model$ contains trapdoors may
use existing backdoor removal methods to identify and remove them.  First,
Liu \etal proposes to remove backdoors by pruning
redundant neurons ({\em neuron pruning})~\cite{liu2018fine}.  As previous work
demonstrates~\cite{wangneural}, normal model accuracy drops rapidly when
pruning redundant neurons. Furthermore, pruning changes the decision
boundaries of the pruned model significantly from those of the
original model.  Hence, adversarial examples  
that fool the pruned model do not transfer well to the original, since
adversarial attacks only transfer between models with 
similarly decision boundaries~\cite{suciu2018does,   demontis2019adversarial}.

We empirically validated this on a pruned 
single-label defended \mnist{}, \gtsrb{}, 
\cifar{}, and \youtube{} models against the six different attacks. We prune
neurons as suggested by~\cite{liu2018fine}. However, we observe that normal
accuracy of the model drops rapidly while pruning ($>32.23\%$ drop). Due
to the significant discrepancy between the pruned and the original
models, adversarial samples crafted on the pruned model do not transfer
to the original trapdoored model. Attack success is $<4.67\%$.  

More recently proposed backdoor defenses ~\cite{wangneural, qiao2019defending,
  liu2019abs} detect backdoors by finding differences between normal and
infected label(s). All of these assume only one or a small number of labels
are infected by backdoors, so that they can be identified as anomalies.
Authors of \textit{Neural Cleanse}~\cite{wangneural} acknowledge that their
approach cannot detect backdoors if more than 36\% of the labels are
infected. Similarly, \cite{qiao2019defending} uses the same technique and has
the same limitations. The authors of ABS~\cite{liu2019abs} explicitly state
that they do not consider multiple backdoors.  We experimentally validate this claim
with \textit{Neural Cleanse} against all-label defended versions of
\mnist{}, \gtsrb{}, \cifar{}, \youtube{}. All the trapdoors in our
trapdoored models avoided detection.

\para{Black-box/Surrogate Model Attacks (Skilled Adversary).} A skilled adversary aware of
trapdoors in $\model$ could use a black-box model stealing
attack~\cite{papernotblackbox}, where they repeatedly query $\model$
with synthetic, generated inputs, and use the classification results to train a local
substitute model. Finally, the adversary generates adversarial examples using
the substitute model and used them to attack $\model$.

Black-box attacks must walk a fine line against trapdoors. To generate
adversarial examples that successfully transfer to $\model$, the attacker
must query $\model$ repeatedly with inputs close to the classification
boundary. Yet doing so means that black-box attacks could also import the
trapdoors of $\model$ into the substitute model.

We test the effectiveness of black box attacks by defending
single label \gtsrb{} models as described in
Section~\ref{sec:single_label_eval}. We construct the substitute model 
following~\cite{papernotblackbox} and use it to generate adversarial attack
images to attack our
original model $\model$. In our tests, we consistently observe that the substitute
model does indeed inherit the trapdoors from $\model$. 
A trapdoored model can reliably detect adversarial examples
generated from black-box substitute models with $>95\%$ success at $5\%$ false
positive rate, for all six attacks (FGSM, PGD, CW, EN, BPDA, SPSA).

If somehow an attacker obtained access to the full training dataset
used by the model and used it to build a surrogate model, they could
reproduce the original clean model.  We consider this possibility later in
this subsection. 

\begin{table}[t]
  \caption{Targeted transferability of Adversarial Examples from a model restored by
    unlearning, to its trapdoored counterpart.}
\label{fig:transfer}
\vspace{-0.1in}
\begin{tabular}{|c|c|c|c|c|c|c|}
\hline
Model        & CW      & EN     & PGD    & BPDA   & SPSA   & FGSM    \\ \hline
GTSRB        & $1.5\%$   & $2.6\%$  & $2.0\%$  & $1.0\%$  & $0.0\%$  & $4.7\%$   \\ \hline
CIFAR10      & $4.4\%$   & $4.4\%$  & $5.6\%$  & $0.0\%$  & $6.7\%$  & $0.0\%$   \\ \hline
Youtube Face & \shawn{$0.0\%$}   & \shawn{$0.0\%$}  & \shawn{$4.1\%$}  & \shawn{$3.3\%$}  & \shawn{$0.0\%$}  & \shawn{$0.0\%$}  \\ \hline
\end{tabular}
\end{table}

\begin{table}[t]
  \caption{Targeted transferability of Adversarial Examples from a model
    trained on clean data to its trapdoored counterpart.}
\label{fig:clean_transfer}
\vspace{-0.1in}
\begin{tabular}{|c|c|c|c|c|c|c|}
\hline
Model        & CW      & EN     & PGD    & BPDA   & SPSA   & FGSM    \\ \hline
GTSRB        & $0.0\%$   & $0.0\%$  & $2.2\%$  & $3.0\%$  & $1.0\%$  & $0.4\%$   \\ \hline
CIFAR10      & $0.0\%$   & $0.0\%$  & $1.7\%$  & $0.7\%$  & $2.8\%$  & $1.2\%$   \\ \hline
Youtube Face & \shawn{$0.0\%$}   & \shawn{$0.0\%$}  & \shawn{$2.1\%$}  & \shawn{$1.7\%$}  & \shawn{$0.0\%$}  & \shawn{$0.0\%$}  \\ \hline
\end{tabular}
\vspace{-0.1in}
\end{table}

\para{Unlearning the Trapdoor (Oracle Adversary).} The goal of this
countermeasure is to completely remove trapdoors from the target model
$\model$ so that attackers can use it to
generate adversarial samples to attack $\model$. Prior work has shown that
adversarial attacks can transfer between models trained on similar
data~\cite{suciu2018does, demontis2019adversarial}. This implies that attacks
may transfer between cleaned and trapdoored versions of the target model.

For this we consider an {\em oracle attacker} who knows everything about a
model's embedded trapdoors, including its exact shape and intensity.
With such knowledge, oracle adversaries seek to construct a
trapdoor-free model by unlearning the trapdoors.

However, we find that such a transfer attack (between $\model$ and a version
of it with the trapdoor unlearned $\model^{unlearn}$) fails. We validate this experimentally
using a single-label defended model. The high level results are summarized in
Table~\ref{fig:transfer}.  We create a new version of each trapdoored model using
backdoor unlearning techniques~\cite{wangneural,cao2018efficient}, which
reduce the trapdoor injection success rate from $99\%$ to negligible rates (around
$2\%$). Unsurprisingly, the trapdoor defense is unable to detect adversarial samples
constructed on the cleaned model $\model^{unlearn}$, with only $7.42\%$ detection success rate
at $5\%$ FPR for \gtsrb{}. However, these undetected adversarial samples do
not transfer to the trapdoored model $\model$. For all six attacks and all
four models, the attack success rate on $\model$ ranges from $0\%$ to $6.7\%$.
We
hypothesize that this might be because a trapdoored model $\model$ must learn unique {\em
  trapdoor distributions} that $\model^{unlearn}$ does not know. This distributional
shift causes significant differences that are enough to prevent adversarial
examples from transferring between models.

\label{subsec:oracle} 
\para{Oracle Access to the Original Clean Model.} Unlearning is unlikely to
precisely recover the original clean model (before the trapdoor).
Finally, we consider the strongest removal attack possible: an oracle
attacker that has somehow obtained access to (or perfectly reproduced) the
original clean model. We evaluate the impact of using the original clean
model to generate adversarial attacks on $\model$.


We are surprised to learn that adding the trapdoor has introduced
significant changes in the original clean model, and has thus destroyed the
transferability of adversarial attacks between them. In
Table~\ref{fig:clean_transfer}, we show the transferability from clean models
to their trapdoored counterparts. For all 6 attacks and all models,
transferability is always never higher than $3\%$. This definitive result
states that no matter how successful an attacker is at removing or unlearning
the trapdoor, or if they otherwise rebuild the original model, their efforts
will fail because adversarial examples from these models do not work on the
trapdoored model $\model$ that is the actual attack target.


\subsection{Advanced Adaptive Attacks (Carlini)}
\label{subsec:evade}

In this section, we present results on two advanced attacks developed in
collaboration with Dr. Nicholas Carlini, both crafted to detect
and leverage weaknesses in the design of trapdoors. Nicholas approached us
after the paper was accepted and offered to test the robustness of trapdoors
by developing more advanced adaptive attacks. Both attacks are significantly
more successful in weakening trapdoor defenses. Here, we describe both attacks,
their key approaches and their results on different types of trapdoor defenses.

\htedit{We note that} a prior version of the paper included results on two other adaptive attacks:
a low learning rate attack that more carefully scans the loss landscape for
adversarial examples, and a lower-bound perturbation attack that tries to
avoid trapdoors by imposing a lower bound on the size of the
perturbation. 
Our results show both attacks
are largely ineffective against trapdoors. Due to space constraints, we focus on two stronger
Carlini attacks here, and refer readers to~\cite{trapdoor-arxiv} for detailed
results on low learning rate and lower-bound perturbation attacks.

\para{Generalities.} Nicholas' two attacks share two general
principles. First, they use different techniques to map out the boundaries of
trapdoors that exist in a protected model, {\em i.e.} their detection signatures, and then
devise ways to compute adversarial perturbations that avoid them.  Second,
they leverage significant compute power, well beyond normal experimental
levels, {\em e.g.} running 10K optimization iterations instead of terminating on
convergence. We consider these quite reasonable for an attacker and do
not consider computational overhead a mitigating factor.

Instead, we evaluaese attacks against variants of trapdoors previously
discussed: randomized neuron signatures (\S\ref{sec:neurons}) and
multiple trapdoors per label (\S\ref{subsec:singlelabel}). First,
randomized neuron signatures can effectively make the signature dynamic at
run time. Since trapdoor avoidance is likely a greedy operation, the
inclusion or exclusion of a small number of neurons can significantly alter
the result. In practice, the defender can quickly switch between different
neuron subsets in unpredictable sequences, making attacker optimizations
difficult. Second, multiple trapdoors cover more of the feature space, making
trapdoor avoidance more difficult. In addition, we hypothesize that
additional trapdoors will remove some natural adversarial examples, much like
a randomized smoothing function. When we evaluate using multiple trapdoors,
we assume the attacker knows exactly the number of trapdoors that have been
inserted into the model per label. Note that we generated 5 models for
each trapdoor configuration to eliminate variance in results due to randomness in trapdoor placement and then performed 10
attacks on each model to generate each single data point.

\begin{figure}[t]
      \centering
      \includegraphics[width=0.75\columnwidth]{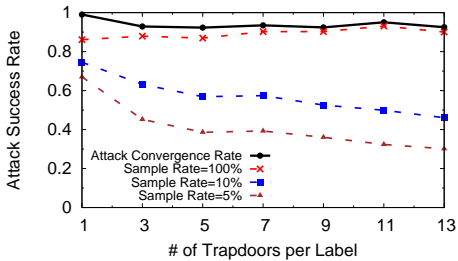}
        \vspace{-0.15in}
        \caption{Oracle Signature Attack success against random neuron
          sampling and multiple trapdoors.}
        \vspace{-0.15in}
    \label{carlini-oracle}
\end{figure}

\begin{figure}[t]
      \centering
      \includegraphics[width=0.75\columnwidth]{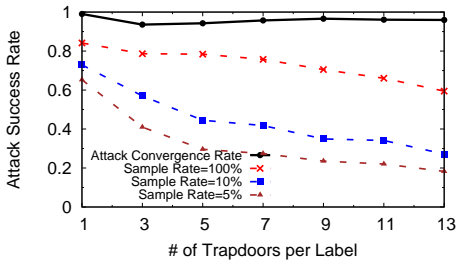}
        \vspace{-0.15in}
        \caption{Trapdoor Vault Attack success against random neuron
          sampling and multiple trapdoors.}
        \vspace{-0.15in}
    \label{vault}
\end{figure}

\para{Oracle Signature Attack.} Nicholas' first attack uses our oracle
adversary assumptions, where the attacker knows the precise values of the
trapdoor. Combined with whitebox access to the model, the attacker is thus
able to generate neuron values matching input with the trapdoor for any layer
in the model. The attacker then jointly optimizes for maximum 
cosine distance from the known trapdoor signature while searching for
adversarial examples.

We plot in Figure~\ref{carlini-oracle} the results of the oracle signature
attack on MNIST with both neuron randomization and multiple trapdoors per
label \htedit{for all label defense}. The top (black) line shows the attack's ability to find successful adversarial
examples. \htedit{The red line marked as} 100\% sample rate represents the attack success after detection by
trapdoor defense \htedit{without neuron randomization}.  Without randomized neuron sampling, the signature attack
succeeds nearly 90\% of the time, effectively breaking the trapdoor
defense. \htedit{The other two lines show the attack success rate
  using neuron randomization.}  We do see that neuron sampling helps to mitigate the attack,
reducing attack success to 40\% after detection, by using only 5\% of neuron
values to build the detection signature. Recall that the defender can update
the random neuron subset in real time with near zero computational cost.  We
note that additional trapdoors have only a slight effect on detection,
because the attacker always knows the number of trapdoors present.

\para{Trapdoor Vault Attack.} Nicholas' second attack, which we call the {\em
  vault attack}, does not require knowledge of the trapdoor
signature. Instead, it estimates the trapdoor signature from the results of
computing adversarial examples on multiple input images.  We extrapolate this
further to target multiple trapdoors by using a histogram/clustering
approach to approximate neuron signatures for each of the $N$ trapdoors. The
attacker then computes joint optimization that maximizes distance to known
trapdoor signatures while searching for adversarial examples. Again we assume
attackers know the exact number of trapdoors present in the model.

We plot in Figure~\ref{vault} the results of the vault attack on MNIST with
both neuron randomization and multiple trapdoors. Again, we see only small
benefits from having multiple trapdoors in the model. However, in this
setting the trapdoor defense does detect
more attacks because of errors in the signature approximation (which can
likely be improved with effort). We do note that when combining randomized
neuron sampling (at 5\%) with multiple trapdoors, we can detect significantly more
attacks, dropping attack success to below 40\%. 

\para{Discussion and Next Steps.} Time constraints greatly limited the
amount of exploration possible in both mitigation mechanisms and further
adaptive attacks. Under base conditions (single trapdoor with 100\% neuron
signature sampling), both attacks effectively break the trapdoor defense. While our
preliminary results show some promise of mitigation, clearly much more work
is needed to explore additional defenses (and more powerful adaptive
attacks).

These attacks are dramatically more effective than other countermeasures
because they were custom-tailored to target trapdoors. We consider their
efficacy as validation that defense papers should work harder to include more
rigorous, targeted adaptive attacks.

\section{Conclusion and Future Work}
\label{sec:discussion}

In this paper, we propose using honeypots to defend DNNs against adversarial
examples. Unlike traditional defenses, our proposed method trains trapdoors into normal models to
introduce controlled vulnerabilities (traps) into the model.  Trapdoors can
defend all labels or particular labels of interest. Across multiple application
domains, our trapdoor-based defense has high detection success against
adversarial examples generated by a suite of state-of-the-art adversarial attacks,
including \cw{}, \elasticnet{}, \pgd{}, BPDA, FGSM, and SPSA, with negligible
impact on normal input classification.

In addition to analytical proofs of the impact of trapdoors on adversarial
attacks, we evaluate and confirm trapdoors' robustness against multiple strong adaptive
attacks, including black-box attacks and unlearning attacks.
Our results on Carlini's oracle and vault attacks show that trapdoors do have
significant vulnerabilities. While randomized neuron signatures help
mitigation, it is clear that further effort is necessary to study 
both stronger attacks and mitigation strategies on honeypot-based defenses.

\section*{Acknowledgments}
We are thankful for significant time and effort contributed by
Nicholas Carlini in helping us develop stronger adaptive attacks on trapdoors. We have learned
much in the process. We also thank our shepherd Ting Wang and anonymous reviewers for their
constructive feedback. This work is supported in part by NSF grants
CNS-1949650, CNS-1923778, CNS-1705042, and by the DARPA GARD program.  Any
opinions, findings, and conclusions or recommendations expressed in this
material are those of the authors and do not necessarily reflect the views of
any funding agencies.

\newpage

\bibliographystyle{ACM-Reference-Format}
\balance
\bibliography{zhao,trapdoor}

\vfill\eject

\section*{Appendix}
\label{sec:appendix}

\subsection{Proof of Theorem 1 \& 2}

\para{Proof of Theorem 1}
\vspace{-0.1in}
\begin{proof}
This theorem assumes that after injecting the trapdoor $\Delta$ into the model, we have 
\begin{equation}
\forall x \in
\mathcal{X}, \;\; \Pr{\model(x+\Delta) =
  y_t \ne \model(x) } \geq 1-\mu.
\label{eq:trapdoor} 
\end{equation}
When an attacker applies gradient-based optimization to find
adversarial perturbations for an input $x$ targeting $y_t$, the above
equation (\ref{eq:trapdoor})    
implies that the partial gradient from $x$
towards $x+\Delta$ becomes the major gradient to achieve 
the target $y_t$. 
Note that $\model(x)$ is the composition of non-linear feature representation $g(x)$ and a linear loss function (e.g. logistic regression):
$\model(x) = g(x) \circ L$ where $L$ represents the linear function.
Therefore, the gradient of $\model(x)$ can be calculated via $g(x)$:
  \begin{align}
 &\frac{\partial ln \model(x)}{\partial x} = \frac{\partial ln [g(x) \circ L] }{\partial x} =
 c \frac{\partial ln g(x) \circ L}{\partial x} 
\label{eq:c4}
\end{align}
Here $c$ is the constant within the linear function $L$. To avoid
ambiguity, we will focus on the derivative on $g(x)$ in the rest of
the proof.

Given (\ref{eq:c4}), we can interpret  (\ref{eq:trapdoor}) in terms of
the major gradient: 
  \begin{equation}
P_{x \in \mathcal{X}}[ \frac{\partial [ln g(x)-ln g(x+\Delta)]}{\partial x}
\geq \eta] \geq 1-\mu,
\label{eq:c0}
\end{equation}
where $\eta$ represents, for the given $x$, the gradient value
required to reach $y_t$ as the classification result. 

Next, since $\forall x \in \mathcal{X}$, $cos(g(A(x)), g(x+\Delta)) \geq \sigma$, and
$\sigma \rightarrow 1$,   without loss of generality we have $g(A(x)) = g(x+\Delta) + \gamma$
where $ |\gamma| << |g(x+\Delta)| $.  Here we rewrite the
adversarial input 
$A(x)$ as $A(x)=x+\epsilon$. Using this condition,  we can prove that
the following two conditions are true.  First, because the value of $\gamma$ does
not depend on $x$, we have
\begin{equation}
\frac{\partial (g(x+\Delta)+\gamma)}{\partial{x}} = \frac{\partial
  g(x+\Delta)}{\partial{x}}.
\label{eq:c1}
\end{equation}
Furthermore, because $|\gamma| << |g(x+\Delta|) $, we have 
\begin{equation}
  \frac{1}{g(x+\Delta) + \gamma} \approx \frac{1}{g(x+\Delta)}.
  \label{eq:c2}
  \end{equation}
Leveraging eq. (\ref{eq:c0})-(\ref{eq:c2}), we have

\begin{align*}
  &P_{x \in \mathcal{X}}[ \frac{\partial [ln g(x)-ln g(x+\epsilon)]}{\partial x}
\geq \eta] \\
  =&P_{x\in \mathcal{X}}[ \frac{1}{g(x)} \frac{\partial g(x)}{\partial
    x} -  \frac{1}{g(x+\epsilon)} \frac{\partial
    {g(x+\epsilon)}}{\partial x}  \geq \eta]  \\
  =&P_{x\in \mathcal{X}}[ \frac{1}{g(x)} \frac{\partial
    g(x)}{\partial x} -  \frac{1}{g(x+\Delta) + \gamma} \frac{\partial (g(x+\Delta)+\gamma)}{\partial{x}}
     \geq \eta] \\
  \approx &P_{x\in \mathcal{X}}[ \frac{1}{g(x)} \frac{\partial
    g(x)}{\partial x} -  \frac{1}{g(x+\Delta)} \frac{\partial (g(x+\Delta))}{\partial{x}}
     \geq \eta] \\
   = &  P_{x \in \mathcal{X}}[ \frac{\partial [ln g(x)-ln g(x+\Delta)]}{\partial x}
       \geq \eta] \\
       \geq & 1-\mu. 
\end{align*}
\vspace{-0.1in}
 \end{proof}

\para{Proof of Theorem 2}
\vspace{-0.1in}
\begin{proof}
This theorem assumes that,  after injecting the trapdoor $\Delta$, we have 
  \begin{equation}
P_{x \in \mathcal{X}_{trap}}[ \frac{\partial [ln g(x)-ln g(x+\Delta)]}{\partial x}
                  \geq \eta] \geq  1-\mu
           \label{eq:cc0}
         \end{equation}

Following the same proof procedure in Theorem 1, we have 
\begin{equation}
         P_{x \in \mathcal{X}_{trap}}[ \frac{\partial [ln g(x)-ln g(x+\epsilon)]}{\partial x}
         \geq \eta] \geq  1-\mu
             \label{eq:cc1}
 \end{equation}

Since $\mathcal{X}_{trap}$ and $\mathcal{X}_{attack}$ are
\term{$\rho$-covert}, 
by definition (see eq. (\ref{eq:rhoconvert})) we have that for any event $C
\subset \Omega$, the largest possible difference between the following 
probabilities $P_{x \in {\mathcal{X}_{attack}}}[C]$ and  $P_{x \in
  {\mathcal{X}_{trap}}}[C]$ is bounded by $\rho$.  

Next let $C$
represent the event: $(\frac{\partial [ln g(x)-ln
                 g(x+\epsilon)]}{\partial x} \geq \eta)$. 
We have, for $x\in \mathcal{X}_{attack}$,
\begin{align*}
&P_{x \in {\mathcal{X}_{attack}}}[ \frac{\partial [ln g(x)-ln
                 g(x+\epsilon)]}{\partial x} \geq \eta] \\
\geq & P_{x \in {\mathcal{X}_{trap}}}[ \frac{\partial [ln g(x)-ln g(x+\epsilon)]}{\partial x} \geq \eta] - \rho \\
  \geq & 1-(\mu+\rho). 
\end{align*}
  \end{proof}
\vspace{-0.1in}

\subsection{Experiment Configuration}

\begin{table}[t]\vspace{-0.1in}
\caption{Model Architecture for MNIST. FC stands for fully-connected layer.}
\label{tab:mnist_cnn}
\centering
\resizebox{1\columnwidth}{!}{
\begin{tabular}{@{}ccccc@{}}
\toprule
Layer Type & \# of Channels & Filter Size & Stride & Activation \\
\midrule
Conv & 16 & 5$\times$5 & 1 & ReLU \\
MaxPool & 16 & 2$\times$2 & 2 & - \\
Conv & 32 & 5$\times$5 & 1 & ReLU \\
MaxPool & 32 & 2$\times$2 & 2 & - \\
FC & 512 & - & - & ReLU \\
FC & 10 & - & - & Softmax \\
\bottomrule
\end{tabular}
}
\end{table}

\begin{figure*}[t]
\centering
\begin{minipage}{0.32\textwidth}
  \centering
  \includegraphics[width=1\textwidth]{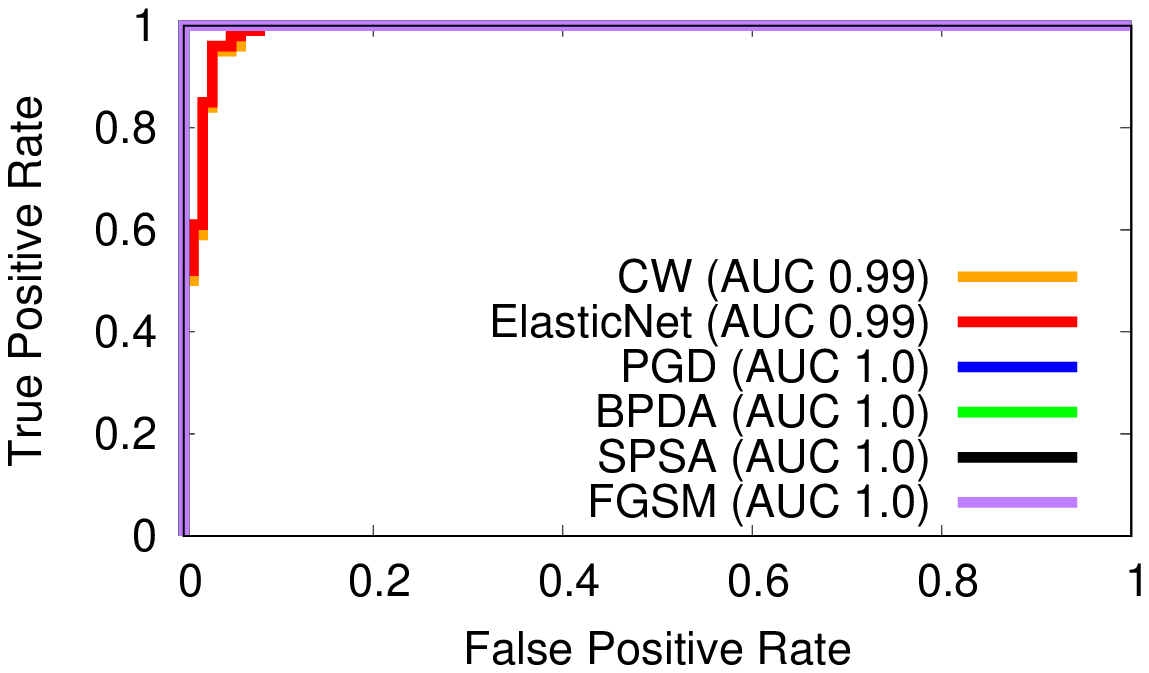}
  \vspace{-0.1in}
  \caption{ROC Curve of detection on \mnist{} with single-label defense.}
  \label{fig:roc_mnist}
\end{minipage}
\hfill
\begin{minipage}{0.32\textwidth}
  \centering
  \includegraphics[width=1\textwidth]{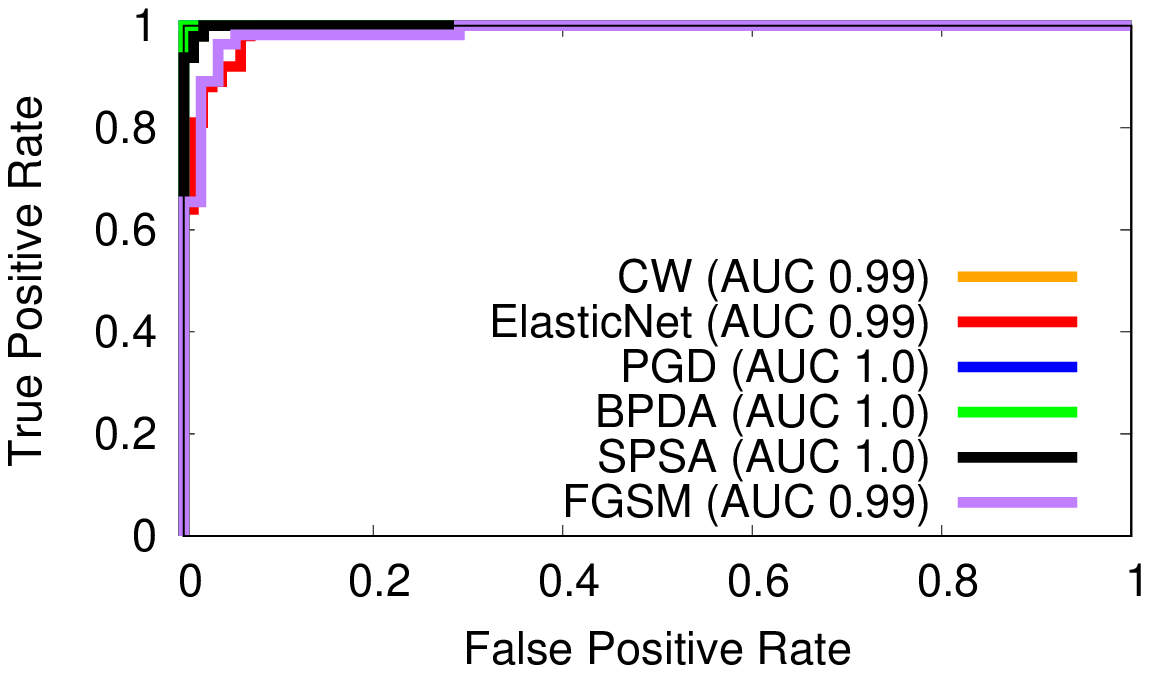}
    \vspace{-0.1in}
  \caption{ROC Curve of detection on \cifar{} with single-label defense.}
  \label{fig:roc_cifar}
\end{minipage}
\hfill
\begin{minipage}{0.32\textwidth}
  \centering
  \includegraphics[width=1\textwidth]{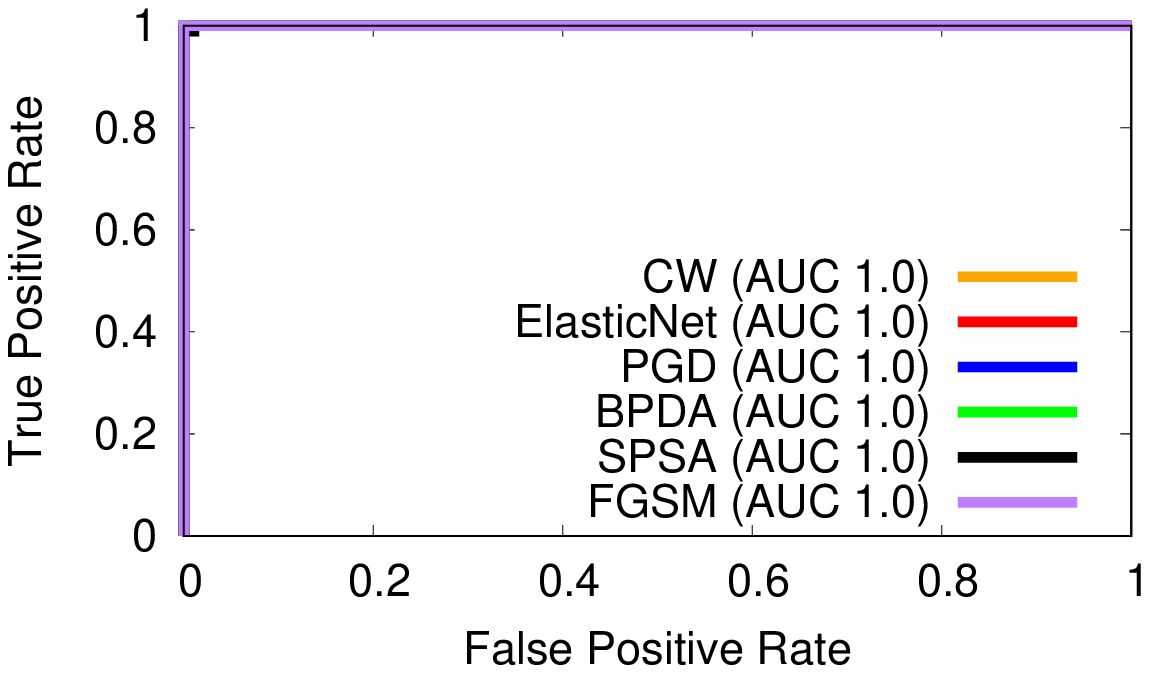}
    \vspace{-0.1in}
  \caption{ROC Curve of detection on \deepid{} with single-label defense.}
  \label{fig:roc_deepid}
\end{minipage}
\end{figure*}

\begin{figure}[t]
  \centering
  \includegraphics[width=0.77\columnwidth]{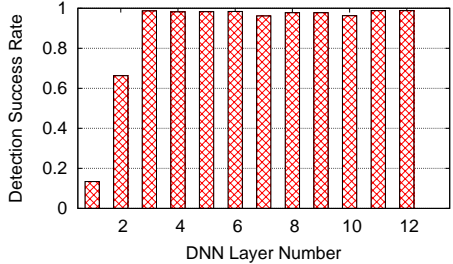}
  \vspace{-0.1in}
  \caption{Detection success rate of \cw{} attack at $5\%$ FPR when using different layers for detection in a \gtsrb{} model.}
  \label{fig:change_layer}
\end{figure}

\begin{table}[t]\vspace{-0.1in}
  \caption{Model Architecture of GTSRB.}
\label{tab:gtsrb_cnn}
\centering
\resizebox{1\columnwidth}{!}{
\begin{tabular}{@{}ccccc@{}}
\hline
Layer Type & \# of Channels & Filter Size & Stride & Activation \\
\hline
Conv & 32 & 3$\times$3 & 1 & ReLU \\
Conv & 32 & 3$\times$3 & 1 & ReLU \\
MaxPool & 32 & 2$\times$2 & 2 & - \\
Conv & 64 & 3$\times$3 & 1 & ReLU \\
Conv & 64 & 3$\times$3 & 1 & ReLU \\
MaxPool & 64 & 2$\times$2 & 2 & - \\
Conv & 128 & 3$\times$3 & 1 & ReLU \\
Conv & 128 & 3$\times$3 & 1 & ReLU \\
MaxPool & 128 & 2$\times$2 & 2 & - \\
FC & 512 & - & - & ReLU \\
FC & 43 & - & - & Softmax \\
\hline
\end{tabular}
}
\end{table}

\begin{table}[t]\vspace{-0.1in}
  \caption{ResNet20 Model Architecture for CIFAR10.}
\label{tab:cifar_cnn}
\resizebox{1\columnwidth}{!}{
\begin{tabular}{@{}cccccc@{}}
\hline
Layer Name (type) & \# of Channels & Activation & Connected to \\
\hline
conv\_1 (Conv) & 16 & ReLU & - \\
conv\_2 (Conv) & 16 & ReLU & conv\_1 \\
conv\_3 (Conv) & 16 & ReLU & pool\_2 \\
conv\_4 (Conv) & 16 & ReLU & conv\_3 \\
conv\_5 (Conv) & 16 & ReLU & conv\_4 \\
conv\_6 (Conv) & 16 & ReLU & conv\_5 \\
conv\_7 (Conv) & 16 & ReLU & conv\_6 \\
conv\_8 (Conv) & 32 & ReLU & conv\_7 \\
conv\_9 (Conv) & 32 & ReLU & conv\_8 \\
conv\_{10} (Conv) & 32 & ReLU & conv\_9 \\
conv\_{11} (Conv) & 32 & ReLU & conv\_{10} \\
conv\_{12} (Conv) & 32 & ReLU & conv\_{11} \\
conv\_{13} (Conv) & 32 & ReLU & conv\_{12} \\
conv\_{14} (Conv) & 32 & ReLU & conv\_{13} \\
conv\_{15} (Conv) & 64 & ReLU & conv\_{14} \\
conv\_{16} (Conv) & 64 & ReLU & conv\_{15} \\
conv\_{17} (Conv) & 64 & ReLU & conv\_{16} \\
conv\_{18} (Conv) & 64 & ReLU & conv\_{17} \\
conv\_{19} (Conv) & 64 & ReLU & conv\_{18} \\
conv\_{20} (Conv) & 64 & ReLU & conv\_{19} \\
conv\_{21} (Conv) & 64 & ReLU & conv\_{20} \\
pool\_1 (AvgPool) & - & - & conv\_{21} \\
dropout\_1 (Dropout) & - & - & pool\_1 \\
fc\_ (FC) & - & Softmax & dropout\_1 \\
\hline
\end{tabular}
}
\end{table}

\begin{table*}[t] 
\centering
\caption{Detailed information on datasets and defense configurations
for each trapdoored model when protecting all labels.}
\label{tab:train_detail}
\centering
\resizebox{0.9\textwidth}{!}{
\begin{tabular}{|c|c|c|c|c|c|c|}
\hline
  \textbf{Model} &
  \begin{tabular}[c]{@{}c@{}}
    \textbf{\#} \\ \textbf{of Labels}
   \end{tabular}  &
\begin{tabular}[c]{@{}c@{}}
\textbf{Training} \\ \textbf{Set Size}
\end{tabular} &
\begin{tabular}[c]{@{}c@{}}
\textbf{Testing} \\ \textbf{Set Size}
\end{tabular} &
\textbf{{Injection Ratio}}
&
\textbf{Mask Ratio} &
\textbf{Training Configuration} \\
\hline
MNIST & 10 & 50,000 & 10,000 & 0.5 & 0.1 & epochs=5, batch=32, optimizer=Adam, lr=0.001 \\
\hline

GTSRB & 43 & 35,288 & 12,630 & 0.5 & 0.1 & epochs=30, batch=32, optimizer=Adam, lr=0.001 \\
\hline

CIFAR10 & 10 & 50,000 & 10,000 & 0.5 & 0.1 & epochs=60, batch=32, optimizer=Adam, lr=0.001 \\
\hline

YouTube Face & 1,283 & 375,645 & 64,150 & 0.5 & 0.2 & epochs=30, batch=32, optimizer=Adam, lr=0.001 \\
\hline
\end{tabular}
}
\end{table*}

\begin{table*}[t]
  \caption{Detailed information on attack configurations. For MNIST experiments, we divid the eps value by 255. }
  \centering
  \label{tab:attackconfig}
  \resizebox{0.85\textwidth}{!}{
  \begin{tabular}{|c|l|}
  \hline
  \textbf{Attack Method} & \multicolumn{1}{c|}{\textbf{Attack Configuration}} \\ \hline
  CW & binary step size = 9, max iterations = 1000, learning rate = 0.05, abort early = True \\ \hline
  PGD & max eps = 8, \# of iteration = 100, eps of each iteration = 0.1 \\ \hline
  ElasticNet & binary step size = 20, max iterations = 1000, learning rate = 0.5, abort early = True \\ \hline
  BPDA & max eps = 8, \# of iteration = 100, eps of each iteration = 0.1 \\ \hline
  SPSA & eps = 8, \# of iteration = 500, learning rate = 0.1 \\ \hline
  FGSM & eps = 8 \\ \hline
  \end{tabular}
  }
\end{table*}

\begin{table}[t]
\centering
\caption{Dataset, complexity, model architecture for each task.}
\vspace{-0.08in}
\label{tab:task_detail}
\resizebox{0.48\textwidth}{!}{
\begin{tabular}{cccccc}
\hline
\multicolumn{1}{c}{\bf Task} & {\bf Dataset}
  & \begin{tabular}[c]{@{}c@{}}{\bf \# of}\\ {\bf
      Labels}\end{tabular} &
                             \multicolumn{1}{c}{\begin{tabular}[c]{@{}c@{}}
                                                  {\bf Input}\\{\bf Size}\end{tabular}}
                               &
                                 \multicolumn{1}{c}{\begin{tabular}[c]{@{}c@{}}{\bf
                                                      Training}\\{\bf Images}\end{tabular}}
                               &
                                 \multicolumn{1}{c}{\begin{tabular}[c]{@{}c@{}}{\bf
                                                      Model}\\{\bf Architecture}\end{tabular}} \\ \hline
\begin{tabular}[c]{@{}c@{}} Digit\\ Recognition\end{tabular} & MNIST & 10 & $28 \times 28 \times 1$ & 60,000 & 2 Conv, 2 Dense [\ref{tab:mnist_cnn}] \\ \hline
\begin{tabular}[c]{@{}c@{}}Traffic Sign\\ Recognition\end{tabular} & GTSRB & 43 &$32 \times 32 \times 3$ & 35,288 & 6 Conv, 2 Dense [\ref{tab:gtsrb_cnn}] \\ \hline
\begin{tabular}[c]{@{}c@{}}Image\\ Recognition\end{tabular} &
                                                              \multicolumn{1}{c}{CIFAR10} & \multicolumn{1}{c}{10} & $32 \times 32 \times 3$ & 50,000 & 20 Resid, 1 Dense [\ref{tab:cifar_cnn}] \\ \hline
\begin{tabular}[c]{@{}c@{}}Facial\\ Recognition\end{tabular} & \multicolumn{1}{c}{\begin{tabular}[c]{@{}c@{}}YouTube\\ Face\end{tabular}} & \multicolumn{1}{c}{1,283} & $224 \times 224 \times 3$ & 375,645 & ResNet-50~\cite{he2016deep} \\ \hline
\end{tabular}
}
\vspace{-0.1in}
\end{table}

\begin{table}[t]
\centering
\caption{Trapdoored model and original model classification
  accuracy when injecting trapdoors for all labels. }
\label{tab:attack_performance}
\vspace{-0.05in}
\resizebox{0.95\columnwidth}{!}{
\begin{tabular}{ccc}
\hline
\textbf{Model}         & \textbf{\begin{tabular}[c]{@{}c@{}}Original Model\\ Classification Accuracy\end{tabular}} & \textbf{\begin{tabular}[c]{@{}c@{}}Trapdoored Model (All Labels) \\ Classification Accuracy \end{tabular}} \\ \hline
\textbf{MNIST}        & 99.2\%                                                                                 & 98.6\%   \\ \hline
\textbf{GTSRB}        & 97.3\%                                                                                 & 96.3\%                                                                                                          \\ \hline
\textbf{CIFAR10}      & 87.3\%                                                                                 & 86.9\%                                                                                                          \\ \hline
\textbf{YouTube Face} & \shawn{99.4\%}                                                                                 & \shawn{98.8\%}                                                                                                          \\ \hline
\end{tabular}}
\vspace{-0.15in}
\end{table}

\para{Evaluation Dataset. } We discuss in details of training datasets we used for the evaluation. 
\begin{packed_itemize}

\item {\em Hand-written Digit Recognition} (\mnist{}) --  
  This task seeks to recognize
  $10$ handwritten digits (0-9) in black and white
  images~\cite{lecun1995mnist}. The dataset consists of $60,000$ training images
  and $10,000$ test images. The DNN model is a standard $4$-layer
  convolutional neural network (see Table~\ref{tab:mnist_cnn}).

\item {\em Traffic Sign Recognition} (\gtsrb{}) --  Here the goal is to recognize $43$
  different traffic signs,  emulating an application for 
  self-driving cars. We use the German Traffic Sign Benchmark
  dataset (GTSRB), which contains $35.3$K colored training images and
  $12.6$K testing images~\cite{gtsrb}. The model consists of $6$ convolution layers and $2$
  dense layers (see Table~\ref{tab:gtsrb_cnn}). This task is 1) commonly used as an adversarial defense evaluation
  benchmark and 2) represents a real-world setting relevant to our defense. 

\item {\em  Image Recognition} (\cifar{}) -- The task is to recognize $10$
  different objects. The dataset contains $50$K colored training
  images and $10$K testing images~\cite{krizhevsky2009cifar}. The
  model is an Residual Neural Network~(RNN) with $20$ residual blocks and $1$
  dense
  layer~\cite{he2016deep}~(Table~\ref{tab:cifar_cnn}). 
We include this task because of its prevalence in general image classification and adversarial
  defense literature.   

\item {\em Face Recognition} (\youtube{}) -- This task is to 
  recognize faces of $1,283$ different people drawn from the YouTube
  videos~\cite{youtubeface}. We build the dataset from~\cite{youtubeface}
  to include $1,283$ labels,
  $375.6$K training images, and $64.2$K testing
  images~\cite{chen2017targeted}. \shawn{We use a large ResNet-50 architecture
  architecture~\cite{he2016deep} with over $25$ million
  parameters.} We include this task
  because it simulates a more complex facial recognition-based security screening
  scenario. Defending against adversarial attack in this setting is
  important. Furthermore, the large set of labels in this task
  allows us to explore the scalability of our trapdoor-enabled detection. 

\end{packed_itemize}

\para{Model Architecture.}
We now present the architecture of DNN models used in our work. 
\begin{packed_itemize}

\item {\bf \mnist{}} (Table~\ref{tab:mnist_cnn}) is a convolutional
  neural network (CNN) consisting of two pairs of
convolutional layers connected by max pooling layers, followed by two
fully connected layers.
\item {\bf \gtsrb{}} (Table~\ref{tab:gtsrb_cnn}) is a
CNN consisting of three pairs of
convolutional layers connected by max pooling layers, followed by two
fully connected layers.

\item {\bf \cifar{}} (Table~\ref{tab:cifar_cnn}) is also a CNN but
includes 
21 sequential convolutional layers, followed by pooling, dropout, and
fully connected layers.

\item {\bf \youtube{}} is the ResNet-50 model trained on the YouTube Face
dataset. It has $50$ residual blocks with over $25$ millions parameters.

\end{packed_itemize}

\para{Detailed information on attack configuration.} We evaluate the trapdoor-enabled detection using
six adversarial attacks: \cw{}, \elasticnet{}, \pgd{}, BPDA, SPSA, and
FGSM (which we have described in Section~\ref{subsec:attack}). Details
about the 
attack configuration are listed in Table~\ref{tab:attackconfig}. 

\begin{figure}[t]
\centering
\begin{minipage}{0.5\textwidth}
  \subfigure[Single Label Defense Trapdoor]{
    \includegraphics[width=0.45\textwidth]{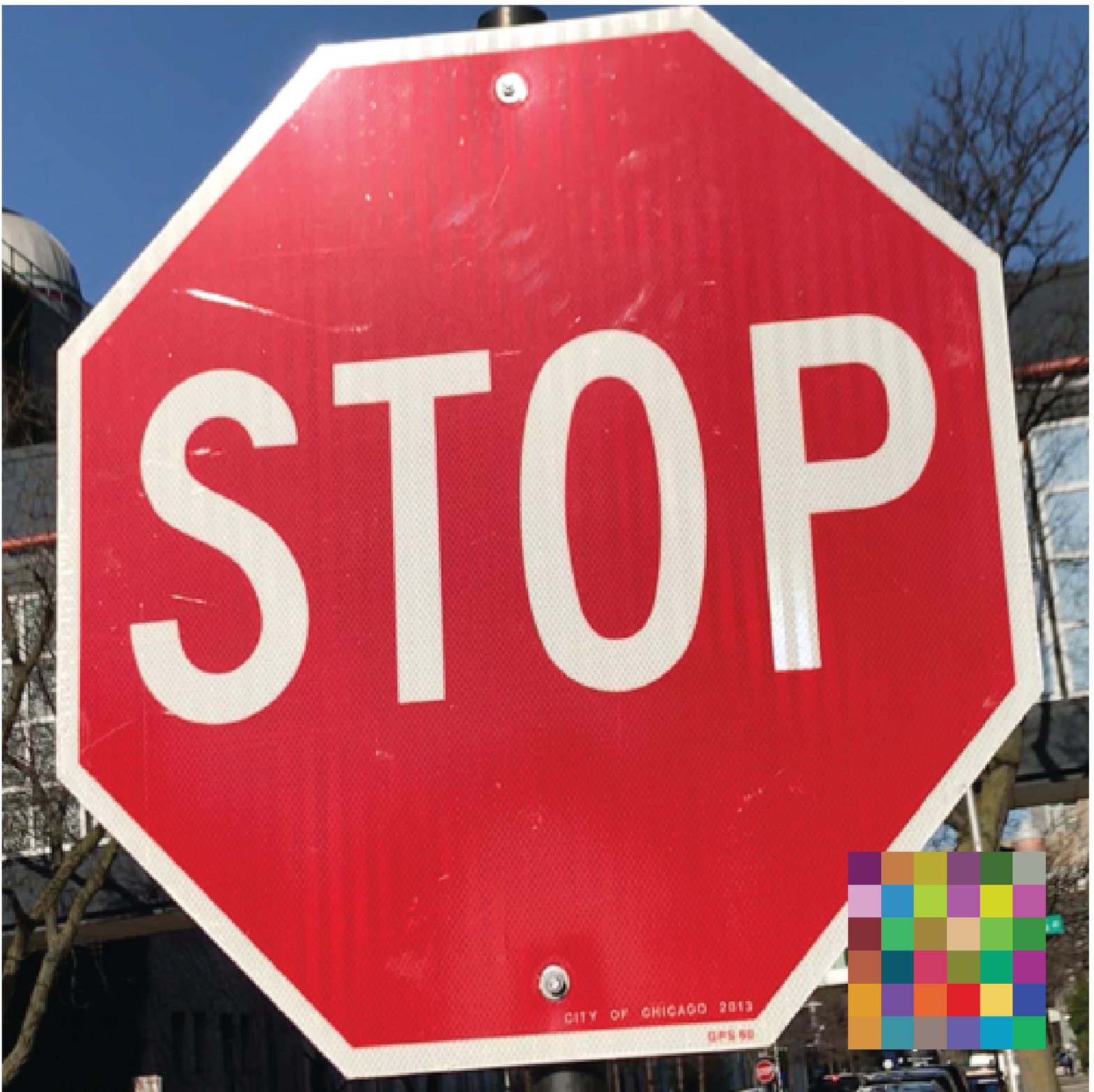}
  } 
  \subfigure[All Label Defense Trapdoor]{
    \includegraphics[width=0.45\textwidth]{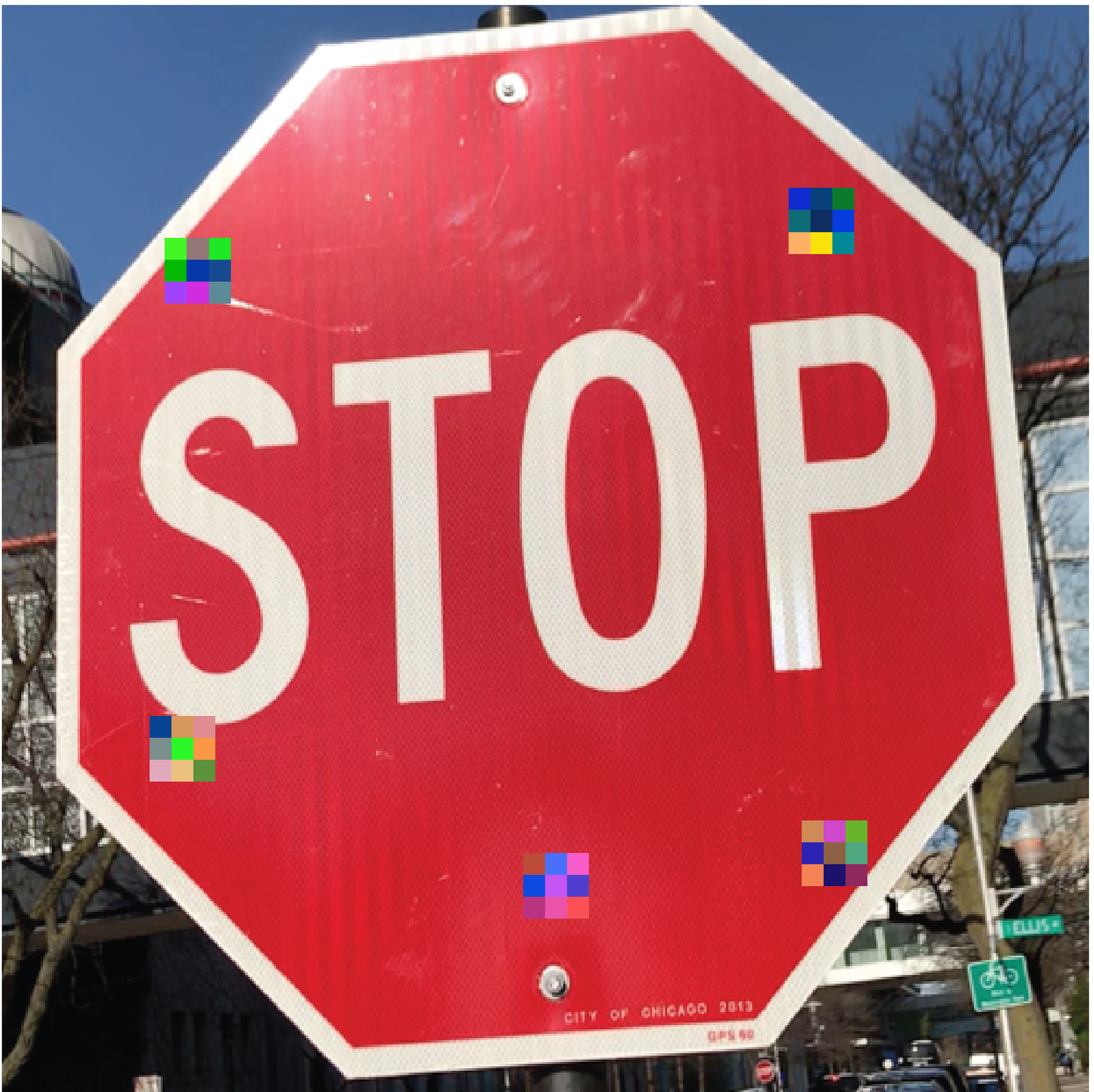}
  }
   \hfill
  \caption{Sample trapdoor examples used in our defense.  While the
    actual trapdoors we used all have a mask ratio of $\kappa=0.1$,
    here we artifically increase $\kappa$ from $0.1$ to $1.0$ in order to highlight the trapdoors
    from the rest of the image content.}
  \label{fig:exampletraps}
\end{minipage}
\end{figure}

\para{Sample Trapdoor Patterns.} Figure~\ref{fig:exampletraps} shows 
sample images that contain a single-label defense trapdoor (a single $6 \times 6$
square) and  that
contain an all-label defense trapdoor (five $3\times 3$ squares).  The
mask ratio of the 
trapdoors used in our experiments is fixed to $\kappa=0.1$.

\para{Datasets and Defense Configuration.}
Tablel~\ref{tab:train_detail} lists the specific datasets and training
process used to inject trapdoors into the four DNN models.

\para{Selecting Trapdoor Injection Ratio.} 
As mentioned earlier, our analysis shows that the size and diversity of the
training data used to inject a trapdoor could affect  its effectivess
of trapping attackers.  To explore this factor, we define {\em
  trapdoor injection ratio} as the ratio between the trapdoored images
and the clean images in the training dataset.  Intuitively, a 
higher injection ratio should allow the model to 
learn the trapdoor better but could potentially degrade normal
classification accuracy. 

We defend the model with different
trapdoor injection ratios and examine the detection success rate.  We
see that only when the injection ratio is very small ({\em e.g.\/} $<0.03$ for
\gtsrb{}),  the model fails to learn the trapdoor and therefore detection
fails.  Otherwise the trapdoor is highly effective in terms of
detecting adversarial examples.  Thus when building the trapdoored
models, we use an  injection ratio of $0.1$ for \mnist{}, \gtsrb{},
\cifar10{}, and $0.01$ for \youtube{} (see
Table~\ref{tab:attackconfig}).

\subsection{Additional Results on Comparing Trapdoor and Adversarial
  Perturbation}

\begin{figure*}[t]
  \centering
  \vspace{-0.2in}
  \subfigure[Trapdoored Model]{
    \includegraphics[width=0.45\textwidth]{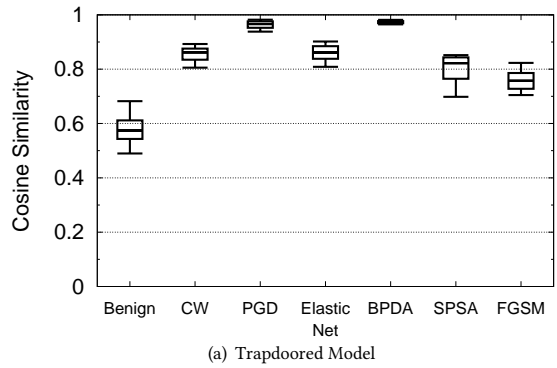}
  }
  \subfigure[Original Model]{
    \includegraphics[width=0.45\textwidth]{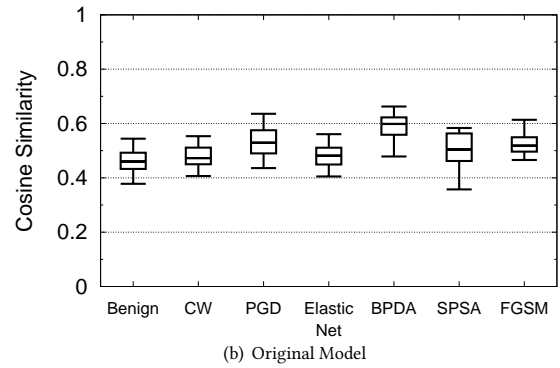}
  }
  \vspace{-0.05in}
  \caption{Comparison of cosine similarity of normal images and
    adversarial images to trapdoored inputs in a trapdoored \cifar{}
    model and in an original (trapdoor-free) \cifar{} model. The boxes show the inter-quartile range, and the whiskers denote the $5^{th}$ and $95^{th}$ percentiles.}
  \label{fig:cosinestat_cifar}
\end{figure*}

\begin{figure*}[t]
  \centering
  \vspace{-0.2in}
  \subfigure[Trapdoored Model]{
    \includegraphics[width=0.45\textwidth]{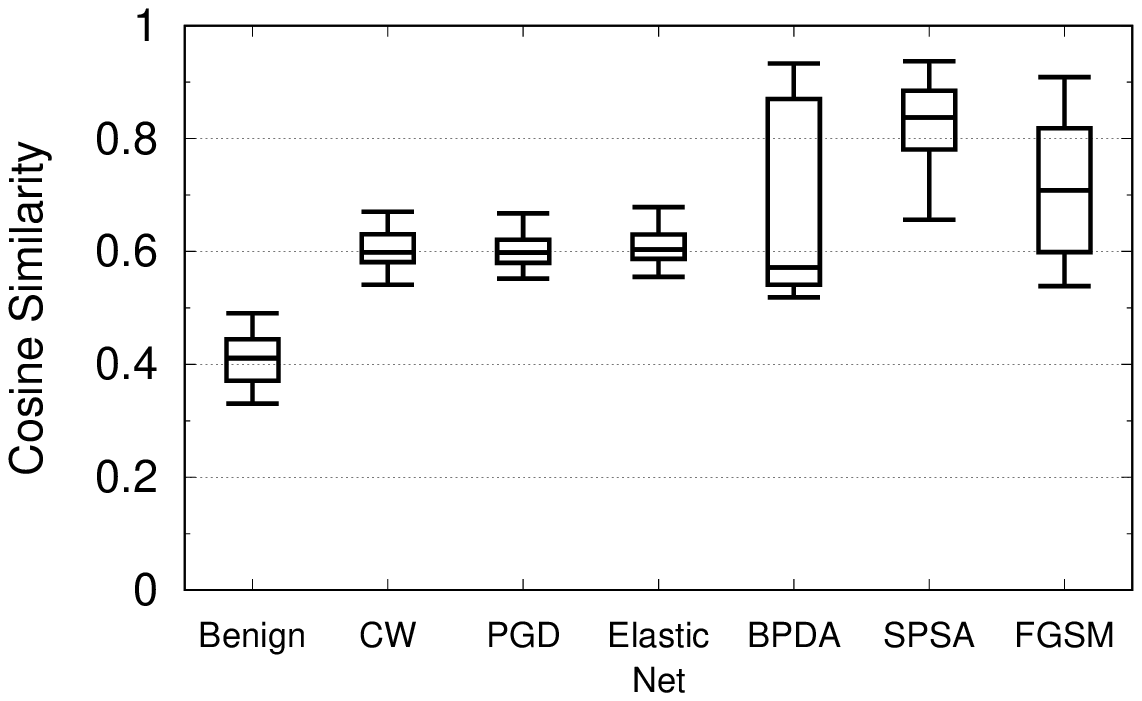}
  }
  \subfigure[Original Model]{
    \includegraphics[width=0.45\textwidth]{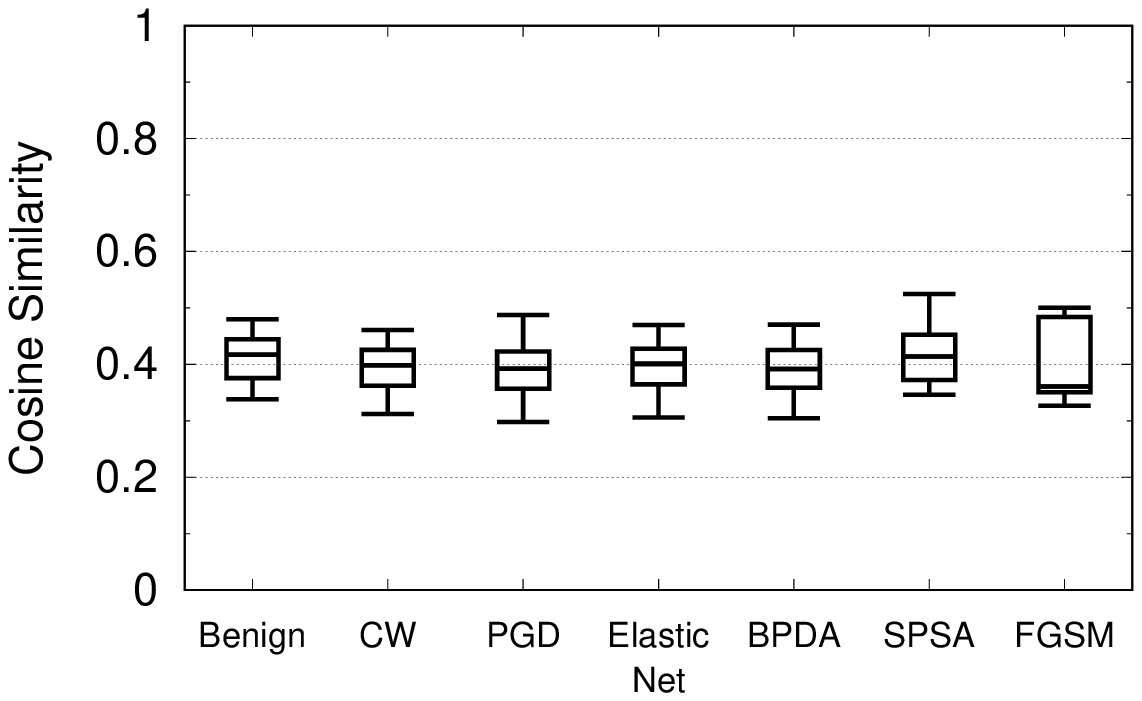}
  }
  \vspace{-0.05in}
  \caption{Comparison of cosine similarity of normal images and
    adversarial images to trapdoored inputs in a trapdoored \youtube{}
    model and in an original (trapdoor-free) \youtube{} model. The boxes show the inter-quartile range, and the whiskers denote the $5^{th}$ and $95^{th}$ percentiles.}
  \label{fig:cosinestat_deepid}
\end{figure*}

Figure~\ref{fig:cosinestat_cifar} and
Figure~\ref{fig:cosinestat_deepid} show that the neuron
signatures of adversarial inputs have high cosine similarity to the
neuron signatures of trapdoors in a trapdoored \cifar{} and \youtube{}
models (left figures),  and the trapdoor-free models (right figures).

\end{document}